\documentclass[preprint,1p,authoryear,12pt]{elsarticle}

\usepackage{amssymb}
\usepackage{multirow}
\usepackage{amsmath}
\usepackage{enumerate}
\usepackage{graphicx}
\usepackage{hyperref}
\usepackage{pdflscape}
\usepackage{longtable}
\usepackage{lscape}
\usepackage{caption}
\usepackage{subcaption}
\usepackage{framed}
\usepackage{picinpar}
\usepackage{enumitem}

\newcommand*{\authorimg}[1]{%
  \raisebox{-.3\baselineskip}{%
    \includegraphics[
      height=6\baselineskip,
      width=6\baselineskip,
      keepaspectratio,
    ]{#1}%
  }%
}

\usepackage{anysize}
\marginsize{3cm}{3cm}{1cm}{3cm}
\usepackage{algorithm,algorithmic}% http://ctan.org/pkg/algorithms

\emergencystretch=\maxdimen
\makeatletter
\newcommand{\ALOOP}[1]{\ALC@it\algorithmicloop\ #1%
  \begin{ALC@loop}}
\newcommand{\ENDALOOP}{\end{ALC@loop}\ALC@it\algorithmicendloop}

\DeclareMathOperator*{\argmaxA}{arg\,max}

\makeatletter
\def\ps@pprintTitle{%
   \let\@oddhead\@empty
   \let\@evenhead\@empty
   \let\@oddfoot\@empty
   \let\@evenfoot\@oddfoot
}
\makeatother

\journal{}

\begin{document}

\begin{frontmatter}


 
\author{Maria Jofre\corref{cor1}}
\ead{maria.jofre@sydney.edu.au}
\cortext[cor1]{Corresponding author}

\author{Richard Gerlach\corref{cor2}}
\ead{richard.gerlach@sydney.edu.au}

\address{Discipline of Business Analytics\\ The University of Sydney Business School, NSW 2006, Australia}

\title{Fighting Accounting Fraud\\ Through Forensic Data Analytics}

%\author[mariaj, richardg]{Maria Jofre, Richard Gerlach}
%\address[mariaj]{The University of Sydney, NSW 2006, Australia}
%\address[richardg]{The University of Sydney, NSW 2006, Australia}

%\author{}
%
%\address{The University of Sydney, NSW 2006, Australia}

\begin{abstract}
Accounting fraud is a global concern representing a significant threat to the financial system stability due to the resulting diminishing of the market confidence and trust of regulatory authorities. Several tricks can be used to commit accounting fraud, hence the need for non-static regulatory interventions that take into account different fraudulent patterns. Accordingly, this study aims to improve the detection of accounting fraud via the implementation of several machine learning methods to better differentiate between fraud and non-fraud companies, and to further assist the task of examination within the riskier firms by evaluating relevant financial indicators. Out-of-sample results suggest there is a great potential in detecting falsified financial statements through statistical modelling and analysis of publicly available accounting information. The proposed methodology can be of assistance to public auditors and regulatory agencies as it facilitates auditing processes, and supports more targeted and effective examinations of accounting reports.

\end{abstract}


\end{frontmatter}


\section{Introduction}
\label{introduction}

In the last few decades, accounting fraud has been drawing a great deal of attention amongst researchers and practitioners, since it is becoming increasingly frequent and diverse. Accounting fraud is one of the most harmful financial crimes as it often results in massive corporate collapses, commonly silenced by powerful high-status executives and managers \citep{Mokhiber2005}. Given their hidden dynamic characteristics, `book cooking' accounting practices are particularly hard to detect, hence the need of more sophisticated tools to assist the exposure of complex fraudulent schemes and the identification of warning signs of manipulated financial reports.

The catastrophic consequences of accounting fraud expose how vulnerable and unprotected the community is in regards to this matter, since most damage is inflicted to investors, employees and government. Several accounting scandals reflect this reality, being the Enron infamous case one of the most controversial. The giant energy company was engaged in a massive fraudulent scheme that culminated abruptly towards the end of 2001 with its impressive collapse and further bankruptcy. Consequently, Enron's investors and stakeholders lost nearly \$74 billion, and 4,500 employees lost their jobs and pensions without proper notice \citep{Swartz2003}. Even though the general opinion describes Enron's failure as unpredictable, \citet{Schilit2010} affirm that the disaster could have been avoided if a careful examination of the public documents during the preceding years of the debacle had been performed. The impressive revenue growth from \$9.2 billion in 1995 to \$100.8 billion in 2000 should have warned the public, especially when considering that profits did not increase at such spectacular rate. They conclude that the use of relevant indicators could be beneficial to further alert the public before a disaster occurs.

In the framework of this study, accounting fraud is defined as the calculated misrepresentation of the financial statement information that is publicly disclosed by companies. The intention is to mislead stakeholders regarding the firm's true financial position, by overstating its expectations on assets, or understating exposure to liabilities; hence the artificial inflation of earnings, as well as its return on equity. Accounting fraud may take the form of either direct manipulation of financial items or via creative methods of accounting \citep{Schilit2010}. Several synonyms of accounting fraud exist in the literature, including the so-called financial statement fraud, corporate fraud and management fraud.

Perpetrators of accounting fraud can be motivated by personal benefit (e.g.: maximisation of compensation packages), or by explicit or implied contractual obligations, such as debt covenants and the need to meet market projections and expected economic growth. The most harm is inflicted to the long-run reputation of the organisation itself, the value destruction of investors and the diminishing of the public's trust in the capital market \citep{Ngai2011}. Other victims often include suppliers, partners, customers, regulatory institutions, enforcement agencies, taxation authorities, the stock exchange, creditors and financial analysts \citep{Pai2011}.

Standard auditing procedures are often insufficient to identify fraudulent accounting reports since most managers recognise the limitations of audits, hence the need for additional dynamic and comprehensive analytical methods to detect accounting fraud accurately and in an early stage \citep{Kaminski2004}. Accordingly, the present study aims to improve the detection rate of accounting fraud offences through the implementation of several machine learning methods and assessment of industry-specific risk indicators, in order to assist the design of an innovative, flexible and responsive corporate regulation tool.

In order to achieve the proposed objective, a thorough forensic data analytic approach is implemented that includes all pertinent steps of a data-driven methodology. The study contributes in the improvement of accounting fraud detection in several ways, including the collection of a comprehensive sample of fraud and non-fraud firms concerning all financial industries, an extensive analysis of financial information and significant differences between genuine and fraudulent reporting, selection of relevant predictors of accounting fraud, contingent analytical modelling of the phenomenon to better recognise fraudulent cases, and identification of financial red-flags as indicators of falsified records.

The rest of the article is organised as follows. A critical review of accounting fraud detection literature is performed in Section \ref{literature} as to summarise commonly used techniques and results achieved in previous studies. Section \ref{methodology} presents a detailed description of the proposed methodology including the studied dataset, sample selection process, explanatory variables examined, variable selection process and machine learning models considered. Section \ref{resultsanddiscussions} illustrates the empirical results of the proposed algorithms and further discussion of key findings. Finally, Section \ref{conclusions} concludes this paper and gives directions for future research.

\section{Accounting Fraud Detection Literature}
\label{literature}

Part of the fraudulent financial reporting literature has focused primarily in the examination of qualitative characteristics related to the board of directors and principal executives, including information of corporate governance structure \citep{Beasley1996, Hansen1996, Bell2000} and insider trading data \citep{Summers1998}. Studies using this kind of information show promising results. However, getting access to such data is very difficult and sometimes even prohibited for most individuals.

On the other hand, studies using publicly available financial statement information are less common and usually incorporate small samples. Generally, the selection of fraud cases is limited to certain conditions and manually matched after with non-fraud observations on the basis of business fundamentals, such as industry, size, maturity, period and more. Undoubtedly, there is an interesting gap in this area of the literature where the selection process of a more representative sample has the potential to be explored and expanded.

With regard to the employed techniques, discriminant analysis and logistic regression are by far the most popular. Such algorithms are commonly considered as a benchmark framework due to their simplicity and low computational cost, and because they have been proven to efficiently detect falsified accounting reporting in relatively small samples \citep{Fanning1998, Spathisetal2002, Kaminski2004, Pai2011}. Better results have been achieved by the implementation of decision trees, a popular machine learning method often used to predict fraudulent accounting records mainly due to their fewer data preparation requirements and intuitive interpretation \citep{Kotsiantis2006, Kirkos2007, Pai2011, Gupta2012, Song2014}.

Alternative and more advanced approaches have also been adopted in order to detect accounting fraud. Neural networks are also in high demand for accounting fraud detection as they have shown promising results when predicting fraudulent reporting practices \citep{Kwon1996, Choi1997, Fanning1998, Feroz2000, Ravisankar2011}. A similar situation is experienced when considering more complex settings, such as support vector machines \citep{Kotsiantis2006, Ravisankar2011, Pai2011, Song2014}, Bayesian networks \citep{Kirkos2007}, genetic programming \citep{Hoogs2007} and hybrid methods \citep{Kotsiantis2006, Song2014}. Nevertheless, the achieved performance of the aforementioned methodologies is counteracted by the considerable drawbacks that these methods entail, including important computational costs and overfitting proneness, as well as struggling when interpreting results \citep{Tu1996, Abe2005}.

A list of prior studies using machine learning techniques for accounting fraud detection is summarised in Table \ref{priorstudies}. Additional methodological details are also provided, such as the size of the chosen samples, number of fraud cases, methods employed and overall accuracy, when available.

\begin{centering}
\begin{table}[htbp]
\caption{Prior studies in detecting accounting fraud}
\vspace{3mm}
\small
\centering
\label{priorstudies}
\scriptsize
\makebox[\textwidth]{%
\begin{tabular}{lcclc}
\hline
\multicolumn{1}{c}{{\bf Study}} & {\bf \begin{tabular}[c]{@{}c@{}}Sample\\ Size\end{tabular}} & {\bf \begin{tabular}[c]{@{}c@{}}Fraud\\ Cases\end{tabular}} & \multicolumn{1}{c}{{\bf Method(s)}} & {\bf \begin{tabular}[c]{@{}c@{}}Overall\\ Accuracy (\%)\end{tabular}} \\ \hline
 & & & & \\
\vspace{0.2cm}
Persons (1995) & 206 & 103 & Logistic Regression & n/a \\
\vspace{0.2cm}
Kwon \& Feroz (1996) & 70 & 35 & \begin{tabular}[c]{@{}l@{}}Neural Networks\\ Logistic Regression\end{tabular} & \begin{tabular}[c]{@{}c@{}}88\\ 47\end{tabular} \\
\vspace{0.2cm}
Choi and Green (1997) & 172 & 86 & Neural Networks & n/a \\
\vspace{0.2cm}
Fanning \& Cogger (1998) & 204 & 102 & \begin{tabular}[c]{@{}l@{}}Logistic Regression\\ Discriminant Analysis\\ Neural Networks\end{tabular} & \begin{tabular}[c]{@{}c@{}}50\\ 52\\ 63\end{tabular} \\
\vspace{0.2cm}
Lee et al. (1999) & 620 & 56 & Logistic Regression & n/a \\
\vspace{0.2cm}
Feroz et al. (2000) & 132 & 42 & \begin{tabular}[c]{@{}l@{}}Neural Networks\\ Logistic Regression\end{tabular} & \begin{tabular}[c]{@{}c@{}}81\\ 70\end{tabular} \\
\vspace{0.2cm}
Spathis (2002) & 76 & 38 & Logistic Regression & 84 \\
\vspace{0.2cm}
Spathis et al. (2002) & 76 & 38 & \begin{tabular}[c]{@{}l@{}}Multicriteria Decision Aid Method\\ Discriminant Analysis\\ Logistic Regression\end{tabular} & \begin{tabular}[c]{@{}c@{}}88\\ 84\\ 81\end{tabular} \\
\vspace{0.2cm}
Lin et al. (2003) & 200 & 40 & \begin{tabular}[c]{@{}l@{}}Neural Networks\\ Logistic Regression\end{tabular} & \begin{tabular}[c]{@{}c@{}}76\\ 79\end{tabular} \\
\vspace{0.2cm}
Kaminski et al. (2004) & 158 & 79 & Discriminant Analysis & n/a \\
\vspace{0.2cm}
Kotsiantis et al. (2006) & 164 & 41 & \begin{tabular}[c]{@{}l@{}}Decision Trees\\ Neural Networks\\ Bayesian Networks\\ Logistic Regression\\ Support Vector Machines\\ Hybrid Decision Support System\end{tabular} & \begin{tabular}[c]{@{}c@{}}91\\ 80\\ 74\\ 75\\ 79\\ 95\end{tabular}   \\
\vspace{0.2cm}
Kirkos et al. (2007) & 76 & 38 & \begin{tabular}[c]{@{}l@{}}Decision Trees\\ Neural Networks\\ Bayesian Networks\end{tabular} & \begin{tabular}[c]{@{}c@{}}74\\ 80\\ 90\end{tabular} \\
\vspace{0.2cm}
Hoogs et al. (2007) & 390 & 51 & Genetic Programming & n/a \\
\vspace{0.2cm}
Lenard et al. (2007) & 30 & 15 & Logistic Regression & 77 \\
\vspace{0.2cm}
Ravisankar et al. (2011) & 202 & 101 & \begin{tabular}[c]{@{}l@{}}Support Vector Machines\\ Genetic Programming\\ Logistic Regression\\ Neural Networks\end{tabular} & \begin{tabular}[c]{@{}c@{}}72\\ 89\\ 71\\ 91\end{tabular} \\
\vspace{0.2cm}
Pai et al. (2011) & 75 & 25 & \begin{tabular}[c]{@{}l@{}}Support Vector Machines\\ Discriminant Analysis\\ Logistic Regression\\ Decision Trees\\ Neural Networks\end{tabular} & \begin{tabular}[c]{@{}c@{}}92\\ 81\\ 79\\ 84\\ 83\end{tabular} \\
\vspace{0.2cm}
Gupta \& Singh (2012) & 114 & 29 & \begin{tabular}[c]{@{}l@{}}Decision Trees\\ Genetic Programming\end{tabular} & \begin{tabular}[c]{@{}c@{}}95\\ 88\end{tabular} \\
\vspace{0.2cm}
Danial et al. (2014) & 130 & 65 & Logistic Regression & 75 \\
\vspace{0.2cm}
Song et al. (2014) & 550 & 110 & \begin{tabular}[c]{@{}l@{}}Logistic Regression\\ Decision Trees\\ Neural Networks\\ Support Vector Machines\end{tabular} & \begin{tabular}[c]{@{}c@{}}78\\ 79\\ 85\\ 86\end{tabular} \\
& & & & \\
\hline
\end{tabular}}
\end{table}
\end{centering}

Many contributions can be attributed to prior studies as all accounting fraud research enhance awareness and knowledge of this phenomenon. Furthermore, it can be said that forensic accounting strongly supports accounting fraud detection and promotes the design of relevant anti-fraud preventive measures.

However, a great deal of work can be further done to improve detection strategies in many ways. First, it can be observed that sample sizes of previous studies are fairly small and that, in general, samples are manually selected. The latter is a highly problematic practice as it is inherently biased and so results cannot be extrapolated to the population. Therefore, increasing the amount of data used to train, validate and test the models is a noticeable enhancement, as well as attempting to collect as many fraudulent cases as possible, and not only the most convenient for the sake of research results.

Moreover, most prior studies focus their analysis in specific industries defined by the Standard Industrial Classification (SIC) system. After careful review, it is surprisingly observed that there are no studies that investigate accounting fraud within financial services firms, situation that can be depicted in Table \ref{studiesindustries}. The main reason for the exclusion of these entities is that they are structurally different and an alternative set of variables may be required since certain financial statement items, such as accounts receivable and inventory, are not available for these companies. Hence ``research to find the variables most useful in the specific industries would be of great value", especially in the poorly examined area of financial services \citep{Fanning1998}. As such, a substantial improvement is achieved in the present study as cases from all industries are included.

Additional improvements in the area of accounting fraud detection can be attained when considering more relevant machine learning methods and performance evaluation metrics. As previously mentioned, complex techniques have been implemented in prior studies, most of them achieving superior performance compared to more basic methods, but the cost of this improvement is relatively high when taking into account the considerable drawbacks that these algorithms entail in terms of computational costs and interpretability. Also, most studies only focus on maximising overall accuracy without further consideration of more suitable assessment measurements.

Consequently, machine learning methods based on decision trees and boosting techniques are implemented in this paper, since their outcome can be very useful when detecting accounting fraud as straightforward classification rules can be extracted, and easily interpreted and replicated by auditors and regulatory agencies. Furthermore, alternative metrics that account for the difference between misclassification costs associated with fraud and non-fraud cases, are proposed to properly measure the predictive ability of the suggested models.

\begin{centering}
\begin{table}[h]
\caption{SIC Industries included in prior studies}
\vspace{3mm}
\small
\centering
\label{studiesindustries}
\footnotesize
\begin{tabular}{ll}
\hline
\multicolumn{1}{c}{\textbf{Study}} & \multicolumn{1}{c}{\textbf{Industry}} \\
\hline
Persons (1995) & Manufacturing and services \\
Kwon \& Feroz (1996) & n/a \\
Choi \& Green (1997) & n/a \\
Fanning \& Cogger (1998) & Financial companies excluded \\
Lee et al. (1999) & Financial companies excluded  \\
Feroz et al. (2000) & Banking companies excluded  \\
Spathis (2002) & Manufacturing firms \\
Spathis et al. (2002) & Manufacturing firms \\
Lin et al. (2003) & n/a \\
Kaminsky (2004) & Banking and insurance firms excluded \\
Kotsiantis et al. (2006) & Manufacturing firms \\
Kirkos et al. (2007) & Manufacturing firms \\
Hoogs et al. (2007) & Financial companies excluded \\
Lenard et al. (2007) & Service-based computer and technology firms \\
Ravisankar et al. (2011) & n/a \\
Pai et al. (2011) & n/a \\
Gupta \& Singh (2012) & n/a \\
Danial et al. (2014) & Financial and insurance sectors excluded \\
Song et al. (2014) & Financial companies excluded \\
\hline
\end{tabular}
\end{table}
\end{centering}

In brief, it can be said that although the proposed techniques of previous studies have increased the detection rate of accounting fraud offences, these are very limited and often not sufficient to uncover complex fraudulent schemes. It is fairly clear, then, the need for improved methodologies that assist the fraud detection task to further discover hidden patterns of falsified financial reports in order to expose them as soon as possible and, therefore, rapidly address recovery strategies and attenuate potential losses.

\section{Methodology}
\label{methodology}

\subsection{Forensic Analytics}
\label{forensicanalytics}

According to \citet{VanVlasselaer2015}, fraud offences are not crimes that happen fortuitously but are carefully planned, concealed and committed. Accounting fraud perpetrators are continuously conceiving new ways to commit their offences and, in consequence, always transforming their fraudulent behaviour, thus the complexity of the accounting fraud phenomenon. This deliberate managerial wrongdoing is particularly hard to detect and predict, since it involves deep knowledge of accounting and legal tricks that are intentionally employed to make documents look genuine and error-free.

Forensic data analysis is concerned with the treatment and examination of financial crime offences, hence the relevance of its use to develop an adequate technique for accounting fraud detection. Therefore, a forensic accounting approach is proposed in order to overcome potential auditing failure and further improve examination of public documents through the recommendation of meaningful analysis of accounting items.

\subsection{Data}
\label{data}

The data collection task is critical in financial crime-related research, since it is very difficult to find sufficient and accurate data for analysis. In addition, and given the highly sensitive nature of the topic, there is a limited amount of relevant journal articles related to accounting fraud detection, and publication of controversial results may be censored or even prohibited \citep{Bolton2002}. Therefore, a compilation of an exhaustive and representative database containing relevant cases of accounting fraud instances is imperative to further design an adequate and integral fraud-detection method.

In this study, accounting fraud cases are identified considering all Accounting Series Releases (ASR) and Accounting and Auditing Enforcement Releases (AAER) issued by the U.S. Securities and Exchange Commission (SEC) between 1990 and 2012. In particular, all public litigation releases involving deceptive reporting were hand-collected first from the SEC's website\footnote{SEC Sanctions Database: \url{https://www.secwhistlebloweradvocate.com/program/sec-enforcement/sanctions-database/}} and then cross-validated with an official accounting fraud database provided by the Securities and Class Action Clearinghouse (SCAC), Stanford Law School. Non-public companies were excluded from this study, since the SEC only has jurisdiction over publicly traded companies.

The selection of the studied period is justified based on data availability and practicality considerations. On the one hand, discovered fraud cases published by the SEC include successful enforcement actions with monetary sanctions exceeding \$1 million announced between July 29, 2002 and present. Accounting fraud cases released by the SEC date from 1990 onwards, hence the selection of the year 1990 as the beginning of the studied period. On the other hand, this study began in the middle of 2013, so including this year would have been erroneous considering that many cases of fraud could have been discovered in the remainder of the year. As such, 2012 is selected as the end year of the studied period.

The resulting fraud database consists of 1,594 fraud-year observations identified by company I.D. and fiscal year of the offence. Table \ref{fraudcasesbyindustry} summarises the number of fraudulent observations obtained after splitting fraud cases into the corresponding years of occurrence, particularly arranged by industry.

\begin{centering}
\begin{table}[h]
\caption{Fraud cases by industry}
\vspace{3mm}
\small
\centering
\label{fraudcasesbyindustry}
\footnotesize
\begin{tabular}{llrr}
\hline
\multicolumn{1}{c}{\textbf{SIC}} & \multicolumn{1}{c}{\textbf{Standard Industrial}} & \multicolumn{1}{c}{\textbf{Fraud}} & \multicolumn{1}{c}{\textbf{Perc}} \\
\multicolumn{1}{c}{\textbf{Codes}} & \multicolumn{1}{c}{\textbf{Classification (SIC)}} & \multicolumn{1}{c}{\textbf{Cases}} & \multicolumn{1}{c}{\textbf{(\%)}} \\
\hline
 & & & \\
0100 - 0999 & Agriculture, Forestry and Fishing & 11 & 0.69 \\
1000 - 1799 & Mining and Construction & 52 & 3.26 \\
2000 - 3999 & Manufacturing & 609 & 38.21 \\
4000 - 4999 & Transportation, Communications, Electric and Gas & 106 & 6.65 \\
5000 - 5999 & Wholesale Trade and Retail Trade & 169 & 10.60 \\
6000 - 6799 & Finance, Insurance and Real Estate & 236 & 14.81 \\
7000 - 8999 & Services & 375 & 23.53 \\
9100 - 9729 & Public Administration & 36 & 2.26 \\
 & & & \\
\hline
 & & \textbf{1,594} & \textbf{100} \\
\hline
\end{tabular}
\end{table}
\end{centering}

\subsection{Sample Selection}
\label{sample_selection}

One of the main characteristics that defines the fraud phenomenon so uniquely is that it is an uncommon activity \citep{VanVlasselaer2015}, particularly in the context of accounting fraud, since only a minority of the recorded cases are actually classified as fraudulent. Learning from these rare events is a very challenging task given the small amount of observations available to train predictive models, hence especially difficult to further discriminate between fraudulent and non-fraudulent instances. As \citet{Cerullo1999} express in regards to this matter, ``unrepresentative sample data or too few data observations will result in a model that poorly estimates or predicts future values".

The class-imbalance problem fully emerges when statistical learning models are applied, because they all opt for a naive strategy of classifying all firms as non-fraudulent. As a consequence, accuracy measures show excellent average performance that only reflect the underlying uneven class distribution. Nevertheless, the methods are totally ineffective in detecting positive cases \citep{Chawla2004}. Therefore, the selection of a more proportionate sample in terms of positive and negative cases is required in order to solve the imbalance problem encountered in this study, and also to enhance the discriminatory power of the proposed statistical models.

A stratifying exercise is conducted according to the target variable \textit{Fraud}, where a pairing exercise is performed to match each fraud observations with a non-fraud observation on the basis of industry and fiscal period. Consequently, the sample selection process occurs in two phases, first dividing the dataset by SIC industry and fiscal year, and then randomly selecting non-fraud instances from each subgroup.

A variety of sampling methods can be employed when dealing with imbalanced datasets, individually or in combination, hence an extensive and interesting analysis could be done to select suitable samples of fraud and non-fraud cases. A more detailed discussion about this topic is addressed in Section \ref{limitationsandfuturework}.

\subsection{Variables}
\label{variables}

A great deal of research studies includes subjective judgment and/or qualitative and non-public information into their models, that are only available to auditors and insiders of the sampled firms. Accounting data, on the other hand, is publicly available for external interested parties, hence whether it can be used to detect falsified reporting is an intriguing question \citep{Persons1995}.

The literature suggests that financial statement information is useful for accounting fraud detection. In particular, it can be seen that ratio analysis is very popular for this end suggesting that a careful reading of financial ratios can reasonably expose symptoms of fraudulent behaviour. As such, ratios are calculated to quantify the relation between two financial items and to subsequently define acceptable legitimate values. Therefore, if a fraudulent activity is taking place, financial ratios associated with manipulated accounts will deviate from the normal behaviour and conveniently exhibit signs of accounting fraud.

There has been an interesting debate about which features should be used for detecting falsified reports, but still no agreement on which ones are best for this end. An in-depth analysis of the most severe accounting scandals occurred in the U.S. in the last few decades \citep{Schilit2010} shows that the most frequent tricks managers employ in order to hide debilitated businesses are commonly associated with the manipulation of earnings and cash flow items.

In this manner, and considering relevant and significant variables resulting from prior research work on the topic, this study identifies 20 financial statement ratios that measure the majority of aspects of a firm's financial performance, including leverage, profitability, liquidity and efficiency.

\vskip 2mm
\textbf{Leverage}
\vskip 0.5mm

One of the most important aspects of a firm is leverage, since it represents the potential return of an investment based on the debt structure of the company. When debt is used to purchase assets, then the value of assets exceeds the borrowing cost, basically because debt interest is tax deductible.

However, this practice comes with greater risks for investors, considering that sometimes firms are not able to pay their debt obligations. In consequence, companies having trouble paying their debts may be tempted to manipulate financial statements in order to meet debt covenants. Therefore, high levels of debt should increase the likelihood of accounting fraud, since it transfers the risk from the firm and its managers to shareholders.

This aspect is measured by the ratios of TLTA (total liabilities to total assets), TLTE (total liabilities to total equity) and LTDTA (long-term debt to total assets).

\vskip 2mm
\textbf{Profitability}
\vskip 0.5mm

Profitability measures are used to estimate the ability of a firm to generate earnings compared to its costs, hence the importance of maintaining these metrics in line with market projections. As consequence, executives may be willing to manipulate earnings-related financial statements in order to cover profitability problems when companies are not performing as expected.

To test whether firms with poorer financial condition are more likely to engage in fraudulent financial reporting, relevant ratios associated with income, expenses and retained earnings will be considered. These ratios are: NITA (net income to total assets), RETA (retained earnings to total assets) and EBITTA (earnings before interest and tax to total assets).

\vskip 2mm
\textbf{Liquidity}
\vskip 0.5mm

Liquidity refers to the ability to which an asset can be converted from an investment to cash. This concept is highly important for businesses and investors, since liquid assets reduce in some extent investing risks by ensuring the capacity of a firm to pay off debts as they come due. Consequently, problems involving liquidity may provide an incentive for managers to commit accounting fraud, hence the need to investigate financial ratios related to the liquid composition of assets, as is the case of working capital and current assets. This aspect is evaluated then by the following ratios: WCTA (working capital to total assets), CATA (current assets to total assets), CACL (current assets to current liabilities) and CHNI (cash to net income).

Many investors have alternatively focused their attention on the company's capability to generate cash from its actual business operations. This aspect however, is usually manipulated since ``companies can exert a great deal of discretion when presenting cash flows" \citep{Schilit2010}. Ergo, the importance of thoroughly analyse cash flow from operations and, in particular, evaluate its relationship with reported earnings. Therefore, the CFFONI ratio (cash flow from operations to net income) is further considered. 

\vskip 2mm
\textbf{Efficiency}
\vskip 0.5mm

Financial efficiency refers to the capacity of producing as much as possible using as few resources as possible. Inefficiency usually involves higher costs, hence resulting in poorer firm's performance, which may motivate managers to misstate financial statements that allow subjective estimations, and therefore, are easier to manipulate. Such is the case of accounts receivable, accounts payable, inventory and cost of good sold, so financial ratios related to these accounts are further selected. This aspect is evaluated by ratios involving the aforementioned items, including RVSA (accounts receivable to total sales), RVTA (accounts receivable to total assets), IVTA (inventory to total assets), IVSA (inventory to total sales), IVCA (inventory to current assets), IVCOGS (inventory to cost of good sold) and PYCOGS (accounts payable to cost of good sold).

Efficiency it also linked to capital turnover, which represents the sales generating power of a firm's assets. In order to maintain the appearance of consistent growth, fraudulent managers may be tempted to manipulate sale-related financial items when dealing with competitive situations. Accordingly, two sale-ratios are considered in order to identify possible fictitious trend in growth, including SATA (total sales to total assets) and SATE (total sales to total equity).

A summary of the aforementioned financial ratios is presented in Table \ref{ratios}, along with the category to which they belong to and their respective calculations.

\begin{centering}
\begin{table}[h]
\caption{Summary of considered financial ratios and calculation}
\vspace{3mm}
\small
\centering
\label{ratios}
\footnotesize
\begin{tabular}{lcl}
\hline
\multicolumn{1}{c}{\textbf{Category}} & \multicolumn{1}{c}{\textbf{Financial Ratio}} & \multicolumn{1}{c}{\textbf{Calculation}} \\ \hline
% & & \\
 & TLTA & Total Liabilities / Total Assets \\
Leverage & TLTE & Total Liabilities / Total Equity \\
\vspace{0.15cm}
 & LTDTA & Long-Term Debt / Total Assets \\ 
% & & \\
\hline
% & & \\
 & NITA & Net Income / Total Assets \\
Profitability & RETA & Retained Earnings / Total Assets \\
\vspace{0.15cm}
 & EBITTA & Earning Before Interest and Tax / Total Assets \\
% & & \\
 \hline
% & & \\
 & WCTA & Working Capital / Total Assets \\
 & CATA & Current Assets / Total Assets \\
Liquidity & CACL & Current Assets / Current Liabilities \\
 & CHNI & Cash / Net Income \\
\vspace{0.15cm}
 & CFFONI & Cash Flow From Operations / Net Income \\
% & & \\
\hline
% & & \\
 & RVSA & Accounts Receivable / Total Sales \\
 & RVTA & Accounts Receivable / Total Assets \\
 & IVSA & Inventory / Total Sales \\
Efficiency & IVTA & Inventory / Total Assets \\
 & IVCA & Inventory / Current Assets \\
 & IVCOGS & Inventory / Cost of Good Sold \\
 & PYCOGS & Accounts Payable / Cost of Good Sold \\
 & SATA & Total Sales / Total Assets \\
\vspace{0.15cm}
 & SATE & Total Sales / Total Equity \\
% & & \\
 \hline
\end{tabular}
\end{table}
\end{centering}

\subsection{Variable Selection}
\label{variable_selection}

Most analytical models implemented to detect fraudulent financial reporting start with numerous variables, out of which only a minority actually contribute to their classification power \citep{Baesens2015}. Thereby, a question of interest to the public is whether fewer explanatory variables can be used in order to achieve similar accuracy rates as those accomplished when using more predictors.

A simple yet very informative univariate analysis is performed in this study in order to evaluate potential differences between financial accounts related to fraudulent and genuine reports, and to further select significant financial ratios that may be suggesting that accounting fraud has been or is being committed.

The so-called Mann-Whitney test is a non-parametric method that is commonly employed for this end due to its ease of use and availability in several advanced statistical software. In simple terms, non-parametric methods refer to statistical techniques that do not make assumptions on the data distribution, hence the reason they are also called distribution-free tests \citep{Hollander2013}. These models are particularly useful when there are clear outliers or extreme observations in the data, as is the case of the studied database.

What it sought with this non-parametric hypothesis testing technique is to test if the distribution of fraud data differs significantly compared to non-fraud data. Therefore, the Mann-Whitney test is performed using the rank of the data, that is, the position of each observation within the sample rather than the value \textit{per se}. In light of this, then it is easy to notice that outliers will have a minimal effect on the test, which makes it very robust in terms of extreme values \citep{Sheskin2003}. 

The following hypotheses are specified for the Mann-Whitney test:

\begin{equation}
\begin{aligned}
H_{0}: \textit{the distribution of both groups are equal} \\
H_{1}: \textit{the distribution of both groups are not equal}
\end{aligned}
\end{equation}

If \textit{p-value} is lower than the 0.05 significance level considered in this paper, then the null hypothesis $H_{0}$ can be rejected, so the evidence favours the alternative, $H_{1}$. Therefore, it can be said that there is a significant difference between the non-fraudulent firms and fraudulent firms with regard to the financial ratio of interest.

It is worth mentioning that when the suggested tests were conducted considering all observations regardless of the SIC industry they belong to, almost all variables revealed to be significant, which does not contribute to the analysis substantially. Moreover, assuming fraudsters behave the same across all sectors is fairly naive, so a more elaborated domain-specific examination is reasonably required.

Consequently, twenty Mann-Whitney tests are performed per industry, one per selected financial ratio. Table \ref{ratiosbyindustry} lists industry-specific significant predictors and the relationship with the dependent variable \textit{Fraud}. Interesting differences between sectors emerge from the performed analysis as some ratios are significant or not depending on the industry the observations belongs to.

On one hand, inventory and retained earnings are relevant predictors in the industries of transportation, communication, electric gas and sanitary service, wholesale trade and retail trade, and services. This may be due to the fact that inventory volumes and retained earnings are easily falsified within the aforementioned sectors. On the other hand, manufacturing companies may be tempted to modify items related to liabilities as well as current assets, while finance, insurance and real estate firms manipulate liabilities and cash flow from operation figures.

\subsection{Correlation Analysis}
\label{correlationanalysis}

A very popular technique, often applied in data analytics, is correlation analysis. This method is used to evaluate possible relationships between numerical variables, which is particularly useful when working with accounting items that inevitably interact with each other due to the composition of a financial statement report.

The correlation coefficient quantifies the direction and strength of the implicit relationship of two variables of interest, and only expresses the association between them, not the causality. Nonetheless, if correlation is found between two variables, then it can be used as an indicator of a potential casual relation.

Kendall correlation coefficients will be used to assess monotonic relationships, that could be linear or not, based-on rank similarity \citep{Kendall1955}. Monotonic relationships occur when one variable increases as well as the other variable, or when one variable increases and the other one decreases. The increase/decrease of the analysed variables could happen at the same rate, which is the case of linear relations, or in a dissimilar proportion, which is the case of non-linear associations.

The Kendall correlation is a non-parametric correlation metric, that is, it makes no assumptions on the distribution of the data. It is said to be a measure of rank correlation in the sense that it calculates the relative position of all observations within one variable (rank position), and then compares them with the ranks obtained within the second variable. If observations from both variables have a similar rank (\textit{concordant} observations), then a high positive correlation will be obtained. Conversely, if ranks are dissimilar (\textit{discordant} observations), then negative correlations are expected.

Kendall coefficients always range between $+1$ and $-1$, and they can be calculated using the \textit{Tau-A} statistic defined as follows:

\begin{equation}
\tau_A = \frac{n_c - n_d}{n(n-1)/2}
\end{equation}

where $n_c$ is the number of concordant pair of observations, $n_d$ is the number of discordant pair of observations, and $n$ is the sample size.

\begin{figure}[h]
\caption{Kendall Correlation Matrix}
\label{kendallcorrelation}
\centering
\includegraphics[scale=0.6]{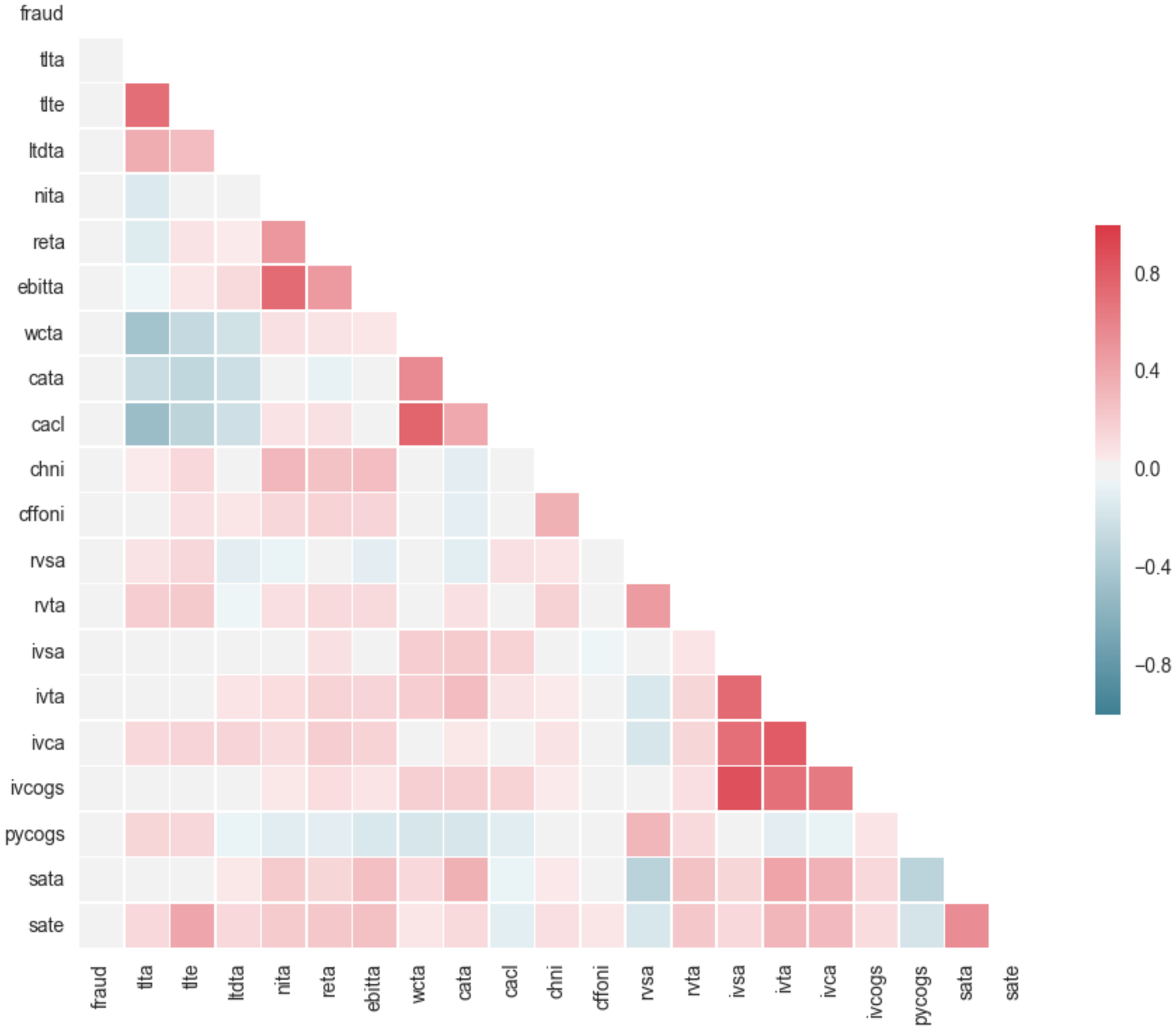}
\end{figure}

The resulting Kendall correlation matrix is presented below (Figure \ref{kendallcorrelation}), summarising the correlation coefficients between all financial ratios. A friendly coloured legend is utilised to facilitate visualisation, where intense red boxes indicate positive relationships and intense blue boxes indicate negative associations.

\begin{landscape}
\begin{table}[]
\centering
\caption{Significant financial ratios by industry}
\vspace{3mm}
\label{ratiosbyindustry}
\scriptsize
\begin{tabular}{lcccccccc}
\hline
\multicolumn{1}{c}{\multirow{2}{*}{\textbf{Ratio*}}} & \textbf{Agriculture,} & \textbf{Mining and} & \multirow{2}{*}{\textbf{Manufacturing}} & \textbf{Transportation,} & \textbf{Wholesale Trade} & \textbf{Finance,} & \multirow{2}{*}{\textbf{Services}} & \textbf{Public} \\
\multicolumn{1}{c}{} & \textbf{Forestry} & \textbf{Construction} & & \textbf{Communications, Electric,} & \textbf{and Retail Trade} & \textbf{Insurance and} & & \textbf{Administration} \\
\multicolumn{1}{c}{} & \textbf{and Fishing} & \textbf{} & \textbf{} & \textbf{Gas and Sanitary Service} & & \textbf{Real Estate} & & \\
\hline
% & & & & & & & & \\
TLTA & & {\normalsize +} & {\normalsize -} & & & {\normalsize +} &  & \\
TLTE & & {\normalsize +} & {\normalsize +} & & & {\normalsize +} &  & \\
LTDTA & & {\normalsize +} & & & & {\normalsize +} &  & {\normalsize +} \\
NITA & & & {\normalsize -} & & & &  & \\
RETA & & {\normalsize +} & {\normalsize -} & {\normalsize +} & {\normalsize +} & {\normalsize +} & {\normalsize +}  & {\normalsize +} \\
EBITTA & {\normalsize +} & {\normalsize +} & {\normalsize +} & & & & {\normalsize +}  & {\normalsize +} \\
WCTA & & & & & & &  & \\
CATA & & & {\normalsize -} & & {\normalsize +} & &  & {\normalsize -} \\
CACL & {\normalsize -} & {\normalsize -} & {\normalsize -} & & & & {\normalsize -}  & {\normalsize -} \\
CHNI & & & & & & {\normalsize -} &  & \\
CFFONI & & & {\normalsize -} & & & &  & \\
RVSA & & {\normalsize -} & {\normalsize -} & & & {\normalsize -} & {\normalsize -}  & \\
RVTA & & {\normalsize +} & & & & &  & \\
IVSA & {\normalsize -} & & & {\normalsize -} & & &  & {\normalsize +} \\
IVTA & & {\normalsize +} & & {\normalsize -} & {\normalsize +} & & {\normalsize -}  & {\normalsize +} \\
IVCA & & {\normalsize +} & {\normalsize +} & {\normalsize -} & {\normalsize +} & & {\normalsize -}  & {\normalsize +} \\
IVCOGS & & {\normalsize +} & & & & {\normalsize +} & {\normalsize -}  & {\normalsize +} \\
PYCOGS & {\normalsize -} & {\normalsize -} & & {\normalsize -} & & {\normalsize -} & {\normalsize -}  & \\
SATA & & {\normalsize +} & & {\normalsize -} & & {\normalsize -} & {\normalsize -}  & {\normalsize -} \\
SATE & & {\normalsize +} & & & & &  & \\
% & & & & & & & & \\
 \hline
\multicolumn{9}{l}{\textit{Notes:}} \\
\multicolumn{9}{l}{{\normalsize+} represents a positive association with the target variable, \textit{Fraud}} \\
\multicolumn{9}{l}{{\normalsize-} represents a negative association with the target variable, \textit{Fraud}} \\
\multicolumn{9}{l}{* Two-tailed test at the 0.05 significance level} \\
\hline
\end{tabular}
\end{table}
\end{landscape}

In addition, a summary of most relevant correlations is shown in Table \ref{kendall_correls}.

\begin{table}[h]
\centering
\caption{Most relevant Kendall correlation coefficients}
\vspace{3mm}
\label{kendall_correls}
\footnotesize
\begin{tabular}{ccr}
\hline
\multicolumn{2}{c}{\textbf{Financial Ratios}} & \textbf{Correlation Coefficient} \\ \hline
 & & \\
IVSA & IVCOGS & 0.8693 \\
IVTA & IVCA & 0.8167 \\
WCTA & CACL & 0.7732 \\
IVSA & IVTA & 0.7485 \\
NITA & RETA & 0.7275 \\
NITA & EBITTA & 0.7275 \\
TLTA & TLTE & 0.7145 \\
IVSA & IVCA & 0.7039 \\
IVCA & IVCOGS & 0.6523 \\
WCTA & CATA & 0.5684 \\
SATA & SATE & 0.5504 \\
 & & \\
 \hline
\end{tabular}
\end{table}

It can be clearly seen that all inventory-related ratios are strongly positive correlated: IVSA, IVTA, IVCA and IVCOGS. Although this situation is completely expected, it entails an important issue when implementing regression models. If two or more variables are highly correlated then multicollinearity emerges, which means some predictors are redundant. As such, the estimated coefficients of the regression model may be inaccurate, and therefore, not very reliable.

Furthermore, a strong positive association has also been found between CACL and WCTA. This is not surprising considering that WC is actually the subtraction of CA and CL, hence a direct relation between these three financial items results from mathematical construction. In addition, and as expected, strong positive correlations between ratios related to profitability have been exposed, which includes both NITA and RETA, as well as NITA and EBITTA. A moderate positive relation between TLTA and TLTE can also be observed, which is completely expected since total assets are calculated as the sum of total liabilities and total equity. Finally, moderate positive correlations have been also exposed between the ratios WCTA and CATA, as well as between SATA and SATE. This latter association makes perfect sense as both ratios are related to sales figures.

Results obtained from the detailed financial ratio analysis performed, which includes non-parametric hypothesis testing and correlation analysis, support the selection of a smaller and meaningful subset of industry-specific explanatory variables. As such, Table \ref{summary_ratios} provided below lists selected financial ratios by SIC industry that will be further utilised for modelling purposes.

\begin{table}[h!]
\centering
\caption{Summary of selected financial ratios by industry domain}
\vspace{3mm}
\label{summary_ratios}
\footnotesize
\begin{tabular}{lcc}
\hline
\multicolumn{1}{c}{\textbf{Industry}} & \multicolumn{1}{c}{\textbf{No. of Ratios}} & \multicolumn{1}{c}{\textbf{Selected Ratios}} \\ \hline
 & & \\
Agriculture, Forestry and Fishing & 4 & RETA, CATA \\
\vspace{0.15cm}
 & & IVSA, PYCOGS \\
\hline
 & & \\
 & & TLTA, TLTE \\
 & & LTDTA, RETA \\
Mining and Construction & 10 & CACL, RVSA \\
 & & IVTA, IVCOGS \\
\vspace{0.15cm}
 & & PYCOGS, SATA \\
\hline
 & & \\
 & & TLTA, TLTE \\
Manufacturing & 6 & RETA, CATA \\
\vspace{0.15cm}
 & & CACL, RVSA \\
\hline
 & & \\
Transportation, Communications, Electric, & 5 & RETA, IVSA \\
Gas and Sanitary Service & & IVTA, SATA \\
\vspace{0.15cm}
 & & PYCOGS \\
\hline
 & & \\
 & & RETA \\
Wholesale Trade and Retail Trade & 3 & CATA \\
\vspace{0.15cm}
 & & IVSA \\
\hline
 & & \\
 & & TLTA, TLTE \\
Finance, Insurance and Real Estate & 8 & LTDTA, RETA \\
 & & CFFONI, IVCOGS \\
\vspace{0.15cm}
 & & PYCOGS, SATA \\
\hline
 & & \\
 & & RETA, CACL \\
Services & 6 & IVSA, IVCOGS \\
\vspace{0.15cm}
 & & PYCOGS, SATA \\
\hline
 & & \\
  & & LTDTA, RETA \\
Public Administration & 8 & CATA, CACL \\
 & & IVSA, IVTA \\
 & & IVCOGS, SATA \\
 & & \\
\hline
\end{tabular}
\end{table}

\newpage
\subsection{Machine Learning Methods}
\label{methods}

The binary outcome model is considered to be the foundational scheme for detecting accounting fraud since the aim is to classify future observations into only two possible values: fraud or non-fraud.

Accordingly, this study assesses the effectiveness of several machine learning models in the identification of fraudulent reporting. First, discriminant analysis and logistic regression are employed as benchmark framework, followed by the implementation of more advanced but easy-to-interpret algorithms, including AdaBoost, decision trees, boosted trees and random forests.

The motivation for using boosting techniques and tree-based methods is supported in part by the poor detection accuracy of basic models and in part by the excessive complexity of more sophisticated approaches, such as neural networks and support vector machines.

In order to achieve a consistent notation throughout the Section, the following conventions are used for mathematical equations:

\begin{itemize}
	\item A superscript $T$ denotes the transpose of a matrix or vector.
	\item $Y=1$: fraudulent observation.
	\item $Y=0$: non-fraudulent observation.
	\item $P(Y=1 \mid X)$: posterior probability of fraud.
	\item $P(Y=0 \mid X)$: posterior probability of non-fraud.
\end{itemize}

It is worth noting that given there are only two possible outcomes, then it holds that:

\begin{equation}
P(Y=0 \mid X) = 1- P(Y=1 \mid X)
\end{equation}

\vskip 2mm

The models were employed as implemented in the Scikit-Learn library \citep{Scikitlearn} and an exhaustive explanation of each algorithm is given in what follows.

\subsubsection{Discriminant Analysis (DA)}

Discriminant analysis is a supervised method used in statistics to address classification problems and to make predictions of a categorical dependent variable. The main idea is to classify an observation into one of the predefined classes using a combination of one or more continuous independent variables in order to generate a discriminant function which best differentiate between the groups.

Subsequently, a decision boundary is generated by fitting class conditional densities $P(X \mid Y)$ to the data using Bayes' rule:

\begin{equation}
P(Y \mid X) = \frac{P(X \mid Y)P(Y)}{P(X)} = \frac{P(X \mid Y)P(Y)}{\sum_y P(X \mid Y = y)P(Y=y)}
\end{equation}

The appropriate class is selected which maximises these conditional probabilities. In the case of accounting fraud, only two classes are of interest; therefore:

\begin{equation}
P(Y=0 \mid X) = \frac{P(X \mid Y=0)P(Y=0)}{P(X \mid Y=0)P(Y=0) + P(X \mid Y=1)P(Y=1)}
\end{equation}

\begin{equation}
P(Y=1 \mid X) = \frac{P(X \mid Y=1)P(Y=1)}{P(X \mid Y=0)P(Y=0) + P(X \mid Y=1)P(Y=1)}
\end{equation}

\vskip 2mm
The optimisation task is ultimately achieved using the training data to estimate class priors, both $P(Y=0)$ and $P(Y=1)$, class means and the covariance matrices. In particular, class priors are estimated as the proportion of instances in each class, that is, number of fraudulent (or non-fraudulent) observation divided by the total number of observations. Class means are estimated using the empirical sample class means. Similarly, covariance matrices are estimated using the empirical sample class covariance matrices.

In accordance with the aforementioned, the following assumptions are made:

\begin{enumerate}
	\item Predictors are all statistically independent.
	\item $P(X \mid Y)$ follows a multivariate Gaussian distribution, with a class-specific mean and covariance matrix.
\end{enumerate}

Different assumptions associated with the covariance matrix will lead to different decision boundaries, one defined by a linear combination of the predictors and another one by a quadratic form.

In both cases, however, the predicted class will be determined using a classification threshold of $0.5$. As such, if the estimated probability of fraud occurrence ($P(Y=1)$) is equal or higher than $0.5$, then the observation will be classified as fraudulent. On the contrary, if $P(Y=1)$ is lower than $0.5$, or equivalently $P(Y=0) \geq 0.5$, then the observation will be classified as non-fraudulent.

\vskip 2mm
\textbf{Linear Discriminant Analysis (LDA)}
\vskip 0.5mm

In the particular case of linear discriminant analysis, a multivariate normal distribution of the predictors is presumed with a distinct mean for each class and a covariance matrix that is common to all classes. For accounting fraud detection, this means that both fraud and non-fraud classes share the same covariance matrix $\Sigma_0 = \Sigma_1 = \Sigma$.

The advantage of a common covariance matrix is that it simplifies the problem by reducing the computational cost of estimating a large number of parameters when the number of predictors is relatively large. Taking this into consideration, then it is true that:

\begin{equation}
P(X \mid Y=0) = \frac{1}{(2\pi)^n\left | \Sigma  \right |^{1/2}}exp\left ( -\frac{1}{2}(X-\mu_0 )^T\Sigma^{-1}(X-\mu_0 ) \right )
\end{equation}

\begin{equation}
P(X \mid Y=1) = \frac{1}{(2\pi)^n\left | \Sigma  \right |^{1/2}}exp\left ( -\frac{1}{2}(X-\mu_1 )^T\Sigma^{-1}(X-\mu_1 ) \right )
\end{equation}

\vskip 2mm
\textbf{Quadratic Discriminant Analysis (QDA)}
\vskip 0.5mm

Furthermore, quadratic discriminant analysis provides a similar approach yet now it is assumed that the covariance matrix is class-specific, i.e.: $X \sim N(\mu_k, \Sigma_k)$ for the $k$th class. Therefore:

\begin{equation}
P(X \mid Y=0) = \frac{1}{(2\pi)^n\left | \Sigma_0  \right |^{1/2}}exp\left ( -\frac{1}{2}(X-\mu_0 )^T\Sigma_0^{-1}(X-\mu_0 ) \right )
\end{equation}

\begin{equation}
P(X \mid Y=1) = \frac{1}{(2\pi)^n\left | \Sigma_1  \right |^{1/2}}exp\left ( -\frac{1}{2}(X-\mu_1 )^T\Sigma_1^{-1}(X-\mu_1 ) \right )
\end{equation}

\subsubsection{Logistic Regression (LR)}

Similar to discriminant analysis, logistic regression is commonly use for performing binary classification. This time the goal is to fit a regression model that estimate the accounting fraud likelihood applying a logistic function that is linear in its argument:

\begin{equation}
\sigma(Z) = \frac{1}{1+exp(-Z)}
\end{equation}

In order to obtain the best classification possible, the posterior probability of belonging to one of both categories is calculated by maximising the likelihood function. Likewise, let $P(Y=1 \mid X)$ be the posterior probability of fraud \citep{Bishop2006}, then:

\begin{equation}
P(Y=1 \mid X) = y(X) = \sigma (w^T X)
\end{equation}

For a dataset \{$x_n,t_n$\}, where $t_n \in$ \{0,1\} and $n=1,...,N$, the likelihood of any specific outcome is given by:

\begin{equation}
P(t \mid w) = \sum_n y_n^{t_n} \{1-y_n\}^{1-t_n}
\end{equation}

where $t = (t_1,...,t_N)^T$ and $y_n = P(Y=1 \mid x_n)$.
\vskip 2mm

As mentioned before, the maximum likelihood estimates of $w$ are obtain by minimising the cross-entropy error function defined by the negative logarithm of the likelihood and then taking its gradient with respect to $w$:

\begin{equation}
E(w) = -ln \{P(t \mid w)\} = -\sum_n \{t_n ln(y_n) + (1 - t_n)ln(1-y_n) \}
\end{equation}

\begin{equation}
\bigtriangledown E(w) = \sum_n (y_n-t_n)x_n
\end{equation}

To finally decide if an observation is classified as fraudulent or non-fraudulent, then a threshold of $0.5$ will be considered. Consequently, if $P(Y=1 \mid X)$ is estimated to be equal or greater than $0.5$, then the observation will be classified as fraudulent. Otherwise, it will be classified as non-fraudulent.

\subsubsection{AdaBoost (AB)}

Adaptive boosting, widely known as AdaBoost, is a machine learning technique used for classification and regression problems that combines multiple 'weak learner' classifiers in order to produce a better boosted classifier. In this context, a weak learner is a function that is only weakly correlated with the response.

The basic idea is to weight observations $w_n$ by how easy or difficult they are to categorise, giving more importance to those that are harder to predict in order to learn from them and further construct better subsequent classifiers. Accordingly, each individual classifier generates an output $G_m(X)$, $m=1,...,M$, for every observation $n$ of the training set. Then, these classifiers are trained on a weighted form using $\alpha_m$ as classifier coefficients. As mentioned before, misclassified instances will be given greater weight when used to train the subsequent classifier \citep{Bishop2006}.

The goal is to minimise a weighted error function $err_m$ in every iteration $m$ taking into account the information and performance of previous classifiers. Ultimately and after the last iteration $M$, a final boost classifier $G(X)$ is constructed as an additive combination of all trained weak learner classifiers $G_m(X)$:

\begin{equation}
G(X) = sign[\sum_m \alpha_m G_m(X)]
\end{equation}

In this case, a classification threshold of $0.5$ has been adopted. As such, an observation will be classified as fraudulent when $G(X)$ is equal or greater than $0.5$, and classified as non-fraudulent when $G(X)$ is lower than $0.5$.

The AdaBoost pseudo code\footnote{Scharth, M. (2017). Statistical Learning and Data Mining, Module 15 [PowerPoint presentation]. Discipline of Business Analytics, The University of Sydney Business School, QBUS6810.} is shown in Algorithm \ref{alg:adaboost}.

\begin{algorithm}[H]
\caption{AdaBoost}
\begin{algorithmic}[1]
\STATE Initialise the observation weights $w_n = 1/N$, $n=1,...,N$.
\FOR{$m=1$ to $M$}
\STATE Fit a classifier $G_m(X)$ to the training data using weights $w_n$.
\STATE Compute the weighted error rate.
\begin{equation*}
err_m = \frac{\sum_n w_n I(y_n \neq G-m(x_n))}{\sum_n w_n}
\end{equation*}
\STATE Compute $\alpha_m=log((1-err_m)/err_m)$.
\STATE Update the weights,
\begin{equation*}
w_n \leftarrow w_n exp[\alpha_m I(y_n \neq G_m(x_n)]
\end{equation*}
\ENDFOR
\STATE Output the classification $G(X) =$ sign[$\sum_m \alpha_m G_m(X)$].
\end{algorithmic}
\label{alg:adaboost}
\end{algorithm}
 
 \subsubsection{Decision Trees (DT)}

Decision trees are a non-parametric supervised learning method that classify observations based on the values of one or more predictors. The advantage of decision trees lies in the straightforward extraction of if-then classification rules easily replicable by auditors and regulatory authorities. Also, no assumptions on the structure of the data is needed, which is very convenient in this case considering the asymmetrical distribution of some explanatory variables.

The structure of a DT consists of nodes representing a test on a particular attribute and branches representing an outcome of the test. The idea is to divide observations into mutually exclusive classes in order to build the smallest set of rules that is consistent with the training data. To identify the attribute that best separates the sample, information gain and entropy reduction are used as estimation criteria.

There are several tree algorithms, such as ID3, C4.5, C5.0 and CART, among others. The chosen method used in this study is the Classification and Regression Trees (CART) characterised by the construction of binary trees based-on feature and threshold selection that provide the largest information gain in each node. This algorithm recursively partitions the space in order to minimise the error or impurity of each node, resulting in terminal nodes that represent homogeneous groups that differ substantially from the others.

Accordingly, let the information at node $m$ be $Q$, then the binary partition of the data is defined by a candidate split $\theta$ that divides the space into two subsets: $Q_{left}(\theta)$ and $Q_{right}(\theta)$.

The error at node $m$ is calculated using an impurity function $H$ evaluated in both partitions, that later is minimised in order to estimate the parameters.

\begin{equation}
G(Q, \theta) = \frac{n_{left}}{N_m} H(Q_{left}(\theta)) + \frac{n_{right}}{N_m} H(Q_{right}(\theta))
\end{equation}

\begin{equation}
\theta^* = argmin_{\theta} G(Q, \theta)
\end{equation}

The impurity function implemented in this study corresponds to the Gini function:

\begin{equation}
H(X_m) = \sum_k p_{mk}(1 - p_{mk})
\end{equation}

where $p_{mk}$ is the proportion of class $k$ observations in node $m$.

It is worth noting that the partitions of the predictor space are based on a greedy algorithm called \textit{recursive binary splitting}. The technique is greedy because at the best split is made at each step of the tree-building process without taking into account the consequences further down the tree. Consequently, in some cases very complex trees are generated as result of this approach. However, a couple of mechanisms can be used in order to avoid this situation, such as setting the minimum number of required observations at a leaf node or setting the maximum depth of the tree.

The tree size is therefore a tuning parameter determining the complexity of the model and it should be selected adaptively from the data. As such, the maximum number of node splits in the current study is settled as 5, optimal valued obtained by cross validation.

Decision trees are remarkably superior than the first two methods used as benchmark - logistic regression and discriminant analysis - considering how easy they are to explain, implement and visualise. Unfortunately, they show some drawbacks that should be mentioned, such as their inherent instability that emerges when little changes in the data cause a large change in the structure of the estimated tree, as well as the lower predictive accuracy when compared to more advanced techniques.

Decision trees can be used as the basic component of powerful prediction methods. Therefore, two additional models that employ decision trees as their foundation, will be introduce in what follows.

\subsubsection{Boosted Trees (BT)}

Similar to AdaBoost, the boosted trees method is an ensemble of weak learners but now in the explicit form of fixed size decision trees as base classifiers.

Accordingly, an iterative process takes place in order to fit a decision tree output $h_m(X)$ in every iteration $m$ to improve the previous model $F_{m}(X)$ by constructing a new model that adds this new information:

\begin{equation}
F_{m+1}(X) = F_m(X) + h_m(X)
\end{equation}

The main idea is to minimise an error function defined by the difference between the old model $F_m(X)$ and the new one $F_{m+1}(X)$, what is called the \textit{residual}, through a gradient boosting algorithm that is much like the gradient descent method used in the logistic regression approach.

In this case, a classification threshold of $0.5$ has also been adopted. In this regard, an observation will be classified as fraudulent when $F_m(X)$ is equal or greater than $0.5$, and classified as non-fraudulent when $F_m(X)$ is lower than $0.5$.

Same as in the decision tree methodology and for consistency, the maximum depth of the fitted trees is established to be 5.

\subsubsection{Random Forests (RF)}

A further enhancement of boosted trees is provided by the random forests approach, one of the most popular bagging techniques. Bootstrap aggregation, or bagging, averages many noisy but approximately unbiased models, which results in a reduction of the variance.

The idea is to fit a classification model to the training data $\mathcal{D}$ to obtain the prediction $\hat{f}(X)$. Bagging averages this prediction over a collection of bootstrap samples\footnote{In statistics, bootstrapping is any test or metric that relies on random sampling with replacement.}. For each bootstrap sample $\mathcal{D}_{b}^\ast$, $b=1,..., B$, the selected classification model is fitted to obtain a prediction $\hat{f}_{b}^\ast(X)$. The bagged classifier selects the class (fraud or non-fraud) with the most ``votes'' from the $B$ classifiers:

\begin{equation}
\hat{y}_{bag}(X) = \argmaxA_c \sum_b I(\hat{y}_{b}^\ast(X)=c)
\end{equation}

Decision trees are ideal candidates for bagging as they capture complex interactions structures in the data, which leads to relatively low biased but high variance. Consequently, classification trees are adopted next for bagging to further construct random forests.

Random forests improve over bagging by adding an adjustment that helps decorrelate the trees. In this context, instead of using all predictors, random forests only select a random subset of the features as split candidates in each step. The rationale behind this methodology is that when establishing a fewer and fixed number of predictors, then more variation in the structure of the model is allowed, which diminishes the correlation between the resulting trees. Interestingly, this new condition makes the average of the fitted trees less variable and therefore more reliable \citep{James2013}.

In building a random forest, $k$ independent variables out of all possible predictors are randomly selected at each node, and later the best split on the considered variables is found. As a last step, all trees are averaged to obtain a final prediction.

The random forests pseudo code\footnote{Scharth, M. (2017). Statistical Learning and Data Mining, Module 13 [PowerPoint presentation]. Discipline of Business Analytics, The University of Sydney Business School, QBUS6810.} is shown in Algorithm \ref{alg:randomforests}.

\begin{algorithm}[H]
\caption{Random Forests}
\begin{algorithmic}[1]
\FOR{$b=1$ to $B$}
\STATE Sample $N$ observations with replacement from the training data $\mathcal{D}$ to obtain the bootstrap sample $\mathcal{D}_{b}^\ast$.
\STATE Grow a random forest tree $T_b$ to $\mathcal{D}_{b}^\ast$ by repeating the following steps for each terminal node of the tree, until the minimum node size is reached:
\STATE (i) Select $k$ variables at random from the $K$ variables.
\STATE (ii) Pick the best variable and split point among the $k$ candidates.
\STATE (iii) Split the node into two daughter nodes.
\ENDFOR
\STATE Output the ensemble of trees $\{T_b\}_{1}^B$.
\end{algorithmic}
\label{alg:randomforests}
\end{algorithm}

In order to be consistent with the previous methodologies, the maximum depth of the estimated trees is established to be 5.

\subsection{Models Assessment}
\label{models_assessment}

An interesting issue related to fraudulent reporting is the difference of misclassification costs. As mentioned previously, most studies only seek to maximise overall accuracy without further analysing more suitable assessment measurements. 

The cost of misclassification differs when dealing with accounting fraud, since a false negative error, which is when a fraud observation is classified as non-fraud, is usually considered more expensive that a false positive error, which is when a non-fraud observation is classified as fraud. The reasoning behind this is that a misclassification of a non-fraud firm may cause an important misuse of resources and time, but a misclassification of a fraudulent company may result in incorrect decisions and economic damage.

Accordingly, the overall accuracy rate is no longer sufficient to assess model performance. Other metrics, such as specificity, sensitivity and precision, are now taken into consideration, as well as G-measure, F-measure and AUC, that are calculated using combinations of these metrics. All mentioned indicators are based on the confusion matrix shown in Table \ref{confusiontable}.

\begin{table}[htbp]
\centering
\caption{Confusion matrix}
\label{confusiontable}
\vskip 3mm
\begin{tabular}{llccc}
\hline
 & & \multicolumn{1}{l}{Predicted Positives} & \multicolumn{1}{l}{} & \multicolumn{1}{l}{Predicted Negatives} \\
\hline
 & & & & \\
 \vspace{0.15cm}
Real Positives &  & TP & & FN \\
Real Negatives &  & FP & & TN \\
 & & & & \\
 \hline
\end{tabular}
\end{table}

Model assessment metrics are described next, including the formula used to calculate them when appropriate.

\begin{enumerate}

\item \textbf{Overall Accuracy}: it measures the ability to differentiate both fraudulent and genuine observations correctly. It is calculated as the proportion of true positive and true negative cases compared to the total number of observations.

\begin{equation}
\small
accuracy = \frac{TP + TN}{TP + FP + FN + TN}
\end{equation}

\item \textbf{Specificity}: it evaluates the ability to determine non-fraudulent cases correctly. As such, it is computed as the proportion of true negative compared to all legitimate negative observations.

\begin{equation}
\small
specificity = \frac{TN}{TN + FP}
\end{equation}

\item \textbf{Sensitivity}: it assesses the capacity to classify fraudulent cases correctly. It is then calculated as the proportion of true positive cases compared to all legitimate positive observations.

\begin{equation}
\small
sensitivity = \frac{TP}{TP + FN}
\end{equation}

\item \textbf{Precision}: it measures the proportion of true positive cases compared to all predicted positive observations.

\begin{equation}
\small
precision = \frac{TP}{TP + FP}
\end{equation}

\item \textbf{G-Mean}: is the geometric mean of sensitivity and specificity measures. As such, it takes into account the ability of correctly classifying both fraudulent and non-fraudulent observations.

\begin{equation}
\small
G-Mean = \sqrt{sensitivity * specificity}
\end{equation}

\item \textbf{F-Measure}: is a metric that integrates both measures of precision and sensitivity

\begin{equation}
\small
F-Measure = \frac{2 * precision * sensitivity}{precision + sensitivity}
\end{equation}

\item \textbf{AUC}: The Area Under the Curve (AUC) is a point estimate of the Receiver Operating Characteristic (ROC) curve, which evaluates the diagnostic ability of a binary classifier model as a function of varying a decision threshold. As such, it assesses both true positive and false positive rates considering different threshold settings. The AUC is the probability that the binary classifier will rank a randomly chosen positive instance higher than a randomly chosen negative one. As such, AUC is always a positive number range between 0 and 1, so the closer to the unit, the better is the model as it means it is correctly separating instances into the non-fraud and fraud groups. The AUC is computed using the trapezoidal rule, which is a commonly used technique for approximating a definite integral.

\end{enumerate}

\vskip 3mm

Regulatory authorities face critical limitations in terms of human resources, budget support and time constrains, thus a detailed investigation of all records and companies is infeasible or too expensive to undertake. Investigations should concentrate on those firms that are more likely to perpetrate accounting fraud. Therefore, it is preferable to focus on models that correctly classify fraudulent observations rather than non-fraudulent cases.

For this reason, G-Mean, F-Measure and AUC will be used as model assessment criteria, since they properly capture both false positive and false negative errors, and mitigate the misclassification issue inherent when detecting accounting fraud offences.

It is worth mentioning, before further presentation and discussion of results, that all classification accuracy metrics are calculated using out-of-sample data, that is, considering all the data points not belonging to the training sample. Furthermore, the considered model will learn the parameters of a prediction function from a subset of the available data and further tested in a different scenario in order to generalise the results. A standard practice in statistics is to hold out part of the dataset, commonly called \textit{testing set}, and use it later to assess the performance of the model.

Therefore, a stratified 10-fold cross-validation approach is implemented before running the proposed variable selection technique. As such, the studied dataset is divided in 10 folds, each one containing an equal number of fraud and non-fraud cases. For each fold, the model is trained by using the remaining nine folds and then validated by using the hold out fold. At last, model performance is calculated as the average performance of all testing folds \citep{Kirkos2007}.

\newpage
\section{Results and Discussions}
\label{resultsanddiscussions}

%\subsection{Results}
%\label{results}

Table \ref{resultsbyind} reports the results of the proposed models by SIC industry.

\begin{table}[H]
\centering
\caption{Prediction accuracy by industry}
\vskip 3mm
\small
\label{resultsbyind}
\begin{tabular}{lccccccc}
\hline
 & Accuracy & Specificity & Sensitivity & Precision & G-Mean & F-Measure & AUC \\
\hline

\multicolumn{8}{l}{\textbf{Agriculture, Forestry and Fishing \textit{(n = 22, p = 4)}}}       \\
LDA & 0.714 & 0.500 & 1.000 & 0.600 & 0.707 & 0.750 & 0.750 \\
QDA & 0.857 & 0.750 & 1.000 & 0.750 & 0.866 & 0.857 & 0.875 \\
LR & 0.714 & 0.500 & 1.000 & 0.600 & 0.707 & 0.750 & 0.750 \\
AB & 0.857 & 0.750 & 1.000 & 0.750 & 0.866 & 0.857 & 0.875 \\
DT & 0.571 & 0.750 & 0.333 & 0.500 & 0.500 & 0.400 & 0.542 \\
BT & 0.571 & 0.750 & 0.333 & 0.500 & 0.500 & 0.400 & 0.542 \\
RF & 0.714 & 0.500 & 1.000 & 0.600 & 0.707 & 0.750 & 0.750 \\
\hline

\multicolumn{8}{l}{\textbf{Mining and Construction \textit{(n = 104, p = 10)}}}       \\
LDA & 0.656 & 0.917 & 0.500 & 0.909 & 0.677 & 0.645 & 0.708 \\
QDA & 0.812 & 0.917 & 0.750 & 0.938 & 0.829 & 0.833 & 0.833 \\
LR & 0.688 & 0.917 & 0.550 & 0.917 & 0.710 & 0.687 & 0.733 \\
AB & 0.625 & 0.667 & 0.600 & 0.750 & 0.632 & 0.667 & 0.633 \\
DT & 0.812 & 0.833 & 0.800 & 0.889 & 0.816 & 0.842 & 0.817 \\
BT & 0.750 & 0.833 & 0.700 & 0.875 & 0.764 & 0.778 & 0.767 \\
RF & 0.781 & 1.000 & 0.650 & 1.000 & 0.806 & 0.788 & 0.825 \\
\hline

\multicolumn{8}{l}{\textbf{Manufacturing \textit{(n = 1,218, p = 6)}}}       \\
LDA & 0.530 & 0.460 & 0.594 & 0.548 & 0.522 & 0.570 & 0.527 \\
QDA & 0.546 & 0.109 & 0.943 & 0.539 & 0.321 & 0.686 & 0.526 \\
LR & 0.530 & 0.425 & 0.625 & 0.545 & 0.516 & 0.583 & 0.525 \\
AB & 0.585 & 0.557 & 0.609 & 0.603 & 0.583 & 0.606 & 0.583 \\
DT & 0.555 & 0.259 & 0.823 & 0.551 & 0.461 & 0.660 & 0.541 \\
BT & 0.574 & 0.621 & 0.531 & 0.607 & 0.574 & 0.567 & 0.576 \\
RF & 0.503 & 0.460 & 0.542 & 0.525 & 0.499 & 0.533 & 0.501 \\
\hline

\multicolumn{8}{l}{\textbf{Transportation, Communications, Electric, Gas and Sanitary Service}}       \\
\multicolumn{8}{l}{\textbf{\textit{(n = 212, p = 5)}}}       \\
LDA & 0.562 & 0.625 & 0.500 & 0.571 & 0.559 & 0.533 & 0.562 \\
QDA & 0.562 & 0.969 & 0.156 & 0.833 & 0.389 & 0.263 & 0.562 \\
LR & 0.578 & 0.594 & 0.562 & 0.581 & 0.578 & 0.571 & 0.578 \\
AB & 0.609 & 0.719 & 0.500 & 0.640 & 0.599 & 0.561 & 0.609 \\
DT & 0.531 & 0.625 & 0.438 & 0.538 & 0.523 & 0.483 & 0.531 \\
BT & 0.672 & 0.719 & 0.625 & 0.690 & 0.670 & 0.656 & 0.672 \\
RF & 0.656 & 0.562 & 0.750 & 0.632 & 0.650 & 0.686 & 0.656 \\
\hline

\multicolumn{8}{l}{\textbf{Wholesale and Retail Trade \textit{(n = 338, p = 3)}}}       \\
LDA & 0.559 & 0.521 & 0.593 & 0.582 & 0.556 & 0.587 & 0.557 \\
QDA & 0.500 & 0.042 & 0.907 & 0.516 & 0.194 & 0.658 & 0.475 \\
LR & 0.549 & 0.521 & 0.574 & 0.574 & 0.547 & 0.574 & 0.547 \\
AB & 0.608 & 0.542 & 0.667 & 0.621 & 0.601 & 0.643 & 0.604 \\
DT & 0.637 & 0.479 & 0.778 & 0.627 & 0.610 & 0.694 & 0.628 \\
BT & 0.745 & 0.771 & 0.722 & 0.780 & 0.746 & 0.750 & 0.747 \\
RF & 0.637 & 0.625 & 0.648 & 0.660 & 0.636 & 0.654 & 0.637 \\
\hline

\end{tabular}
\end{table}

\begin{table}[H]
\centering
\small
\label{table2}
\begin{tabular}{lccccccc}
\hline
 & Accuracy & Specificity & Sensitivity & Precision & G-Mean & F-Measure & AUC \\
\hline

\multicolumn{8}{l}{\textbf{Finance, Insurance and Real Estate \textit{(n = 472, p = 8)}}}       \\
LDA & 0.570 & 0.621 & 0.526 & 0.615 & 0.572 & 0.567 & 0.574 \\
QDA & 0.592 & 0.273 & 0.868 & 0.579 & 0.487 & 0.695 & 0.571 \\
LR & 0.570 & 0.591 & 0.553 & 0.609 & 0.571 & 0.579 & 0.572 \\
AB & 0.648 & 0.621 & 0.671 & 0.671 & 0.646 & 0.671 & 0.646 \\
DT & 0.627 & 0.561 & 0.684 & 0.642 & 0.619 & 0.662 & 0.622 \\
BT & 0.655 & 0.682 & 0.632 & 0.696 & 0.656 & 0.662 & 0.657 \\
RF & 0.627 & 0.652 & 0.605 & 0.667 & 0.628 & 0.634 & 0.628 \\
\hline

\multicolumn{8}{l}{\textbf{Services \textit{(n = 750, p = 6)}}}       \\
LDA & 0.587 & 0.468 & 0.698 & 0.583 & 0.572 & 0.635 & 0.583 \\
QDA & 0.587 & 0.229 & 0.922 & 0.560 & 0.460 & 0.697 & 0.576 \\
LR & 0.587 & 0.495 & 0.672 & 0.586 & 0.577 & 0.627 & 0.584 \\
AB & 0.627 & 0.550 & 0.698 & 0.623 & 0.620 & 0.659 & 0.624 \\
DT & 0.631 & 0.615 & 0.647 & 0.641 & 0.630 & 0.644 & 0.631 \\
BT & 0.631 & 0.550 & 0.707 & 0.626 & 0.624 & 0.664 & 0.629 \\
RF & 0.618 & 0.477 & 0.750 & 0.604 & 0.598 & 0.669 & 0.614 \\
\hline

\multicolumn{8}{l}{\textbf{Public Administration \textit{(n = 72, p = 8)}}}       \\
LDA & 0.636 & 0.400 & 0.833 & 0.625 & 0.577 & 0.714 & 0.617 \\
QDA & 0.818 & 0.900 & 0.750 & 0.900 & 0.822 & 0.818 & 0.825 \\
LR & 0.727 & 0.600 & 0.833 & 0.714 & 0.707 & 0.769 & 0.717 \\
AB & 0.727 & 0.700 & 0.750 & 0.750 & 0.725 & 0.750 & 0.725 \\
DT & 0.773 & 0.700 & 0.833 & 0.769 & 0.764 & 0.800 & 0.767 \\
BT & 0.773 & 0.700 & 0.833 & 0.769 & 0.764 & 0.800 & 0.767 \\
RF & 0.864 & 0.900 & 0.833 & 0.909 & 0.866 & 0.870 & 0.867 \\
\hline

\end{tabular}
\end{table}

It can be seen that results are dissimilar across different industries and machine learning techniques. Best performance of the proposed models is obtained for firms belonging to the Agriculture, Forestry and Fishing industry, Mining and Construction, and Public Administration. Moderate predictive accuracy is achieved in the industries of Wholesale and Retail Trade, Transportation and Communications, and Financials. Inferior accuracy can be observed for Manufacturing and Services industries.

\vskip 2mm
\textit{Agriculture, Forestry and Fishing}
\vskip 0.5mm

In particular, good classification performance is achieved in the industry of Agriculture, Forestry and Fishing probably due to the relatively small size of the sample at issue. Four financial ratios have been considered for modelling purposes, including RETA, CATA, IVSA and PYCOGS. The results indicate that quadratic discriminant analysis and boosted trees are the most accurate models as both achieved an AUC of 0.875. In both cases, 75\% of non-fraud cases are correctly identified, as well as 100\% of fraud cases.

Again, special case must be taken when generalising these results, as a fairly small sample is being considered. It is worth mentioning that no relevant patterns have been found within this industry when constructing a decision tree. Because of the small amount of available data, it was unfeasible to find significant red-flags in this domain.

\vskip 2mm
\textit{Mining and Construction}
\vskip 0.5mm

Good results can also be observed in the case of the Mining and Construction industry, where ten financial ratios were considered as predictors and a relatively big sample has been considered. In this case, superior performance has been achieved by QDA and random forests, mainly because of their remarkable accuracy when predicting negative cases, that is, high values of specificity. Nevertheless, good specificity and sensitivity rates are attained when using decision trees as they correctly classify 83.3\% of non-fraud cases and 80\% of fraud cases.

\begin{figure}[H]
\caption{Decision Tree Visualisation \\ Industry: Mining and Construction}
\label{DT_SIC2}
\centering
\includegraphics[scale=0.39]{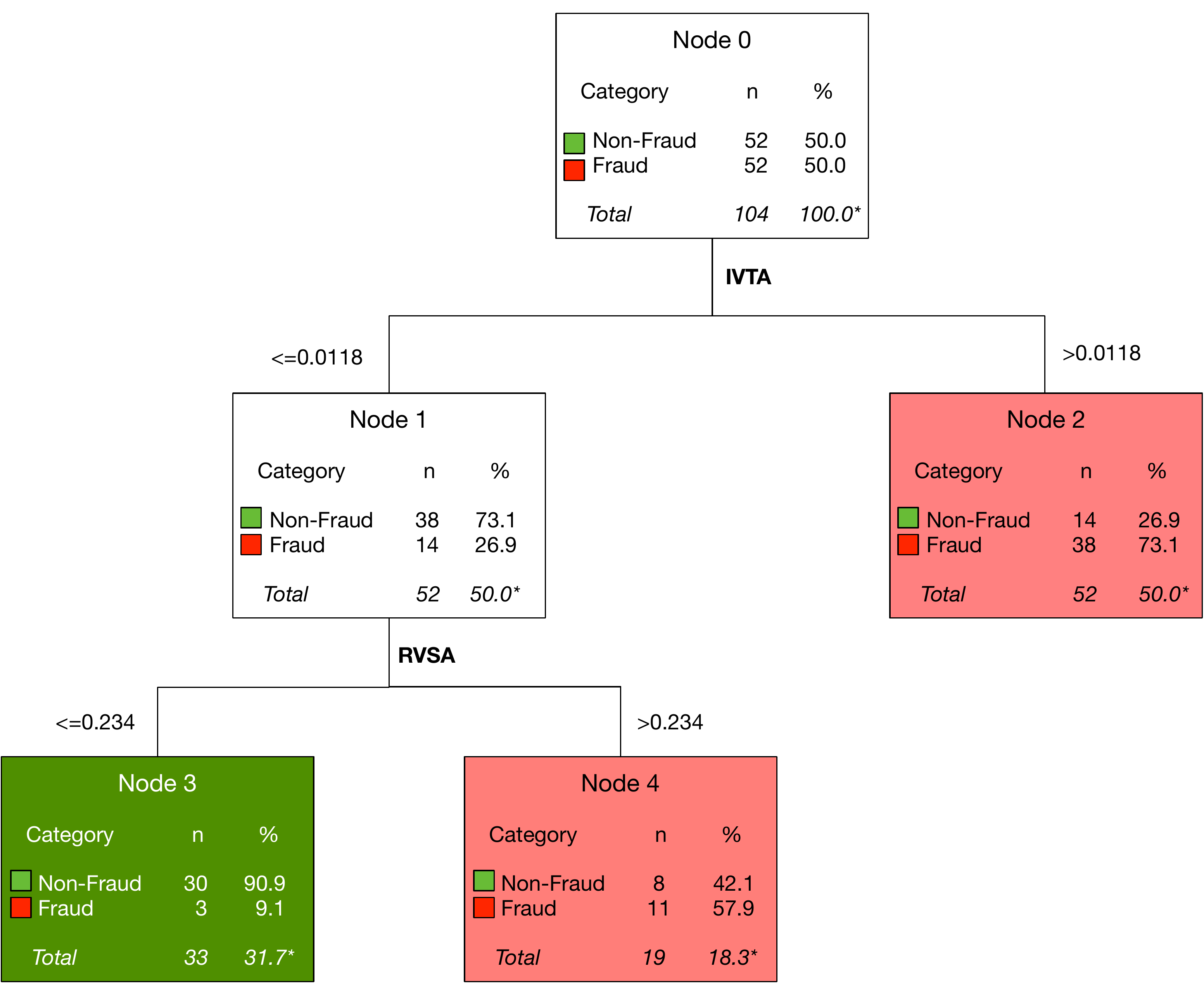}
\end{figure}

More interesting results can be seen when using all observation to construct a decision tree model. As depicted in Figure \ref{DT_SIC2}, two main red-flags, associated with the items of inventory and accounts receivable, can be used to detect fraudulent companies in the Mining and Construction industry. The first one is IVTA, as the evidence suggests that it is more likely to be in presence of fraud when this ratio is bigger than 0.0118, which indicates that fraudulent firms tend to exaggerate inventory levels in this particular industry. Hence, fraud alarms should be activated when inventories represent more than 1.2\% of total assets in mining and construction firms.

The second indicator than can be used to expose falsified reports is RVSA. As such, when inventory levels compared to assets (IVTA) are within the non-fraudulent range (i.e.: lower than 0.0118), then auditors should check if RVSA levels are higher than 0.234. Therefore, the greater the probability of accounting fraud when figures of receivables represent more than 23.4\% of total sales.

\vskip 2mm
\textit{Manufacturing}
\vskip 0.5mm

Inferior performance of all predictive models is achieved when dealing with manufacturing firms. Relatively better results are obtained by boosting techniques. In particular, AdaBoost correctly classifies 55.7\% of non-fraud cases and 60.9\% of fraud cases, which is only a small improve as opposed to random guessing. This is at least surprising as the size of the sample considered is relatively big and predictors have shown significance differences between the groups.

\begin{figure}[H]
\caption{Decision Tree Visualisation \\ Industry: Manufacturing}
\label{DT_SIC3}
\centering
\includegraphics[scale=0.45]{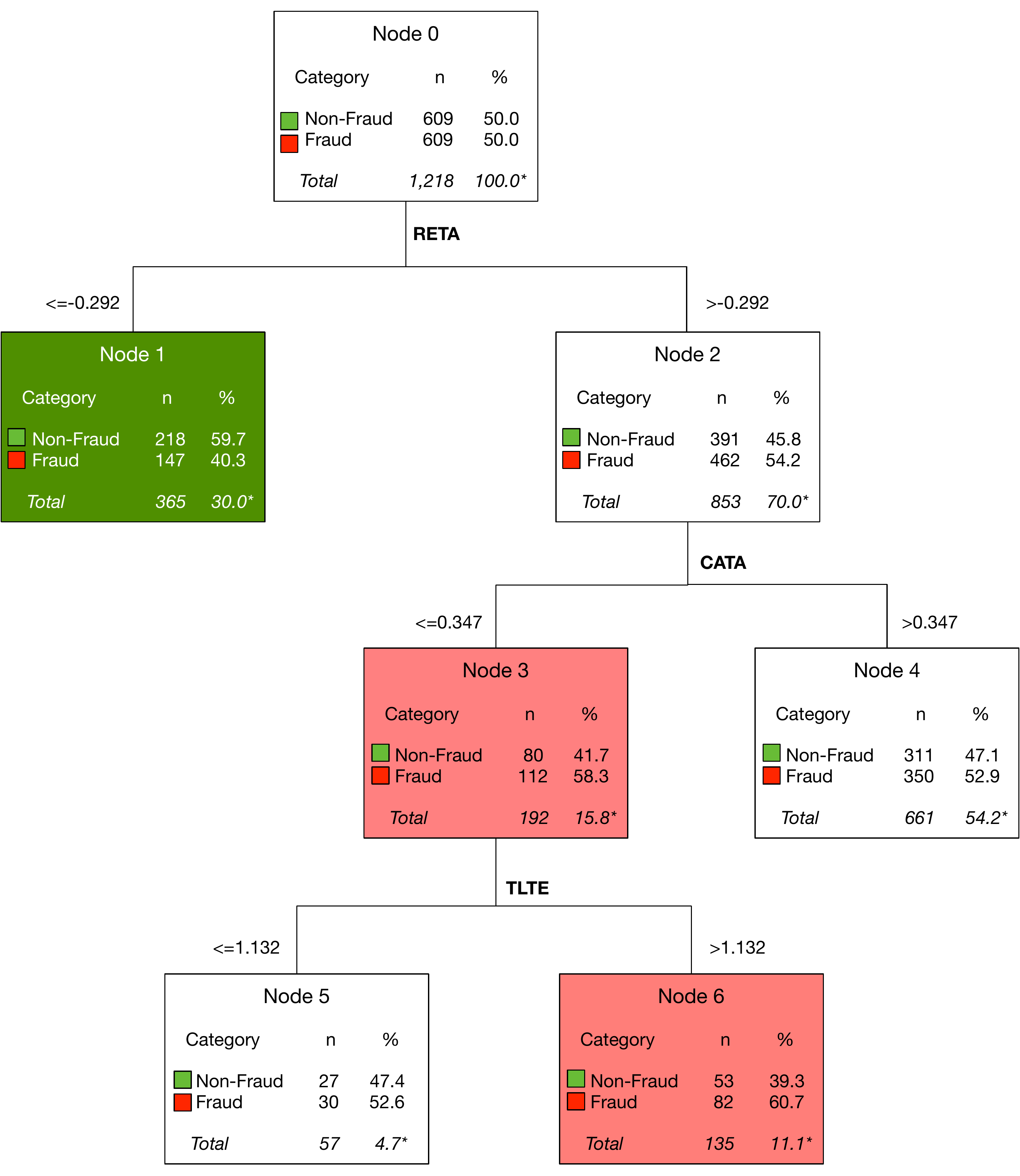}
\end{figure}

The reason for a poor predictive performance can be associated with the complexity of the fraud schemes perpetrated within this industry. Although models show bad performance in general, interesting patterns emerge when implementing a decision tree method using all observations, as it can be seen in Figure \ref{DT_SIC3}. Falsifying reports in this case, usually involve the manipulation of three financial items, that is, retained earnings, current assets and total liabilities.

Moreover, decision tree results indicate that auditors should be more sceptical if RETA is higher that -0.292, CATA lower than 0.347 and TLTE higher than 1.132, since these three red-flags together are often seen when fraud is being committed in manufacturing firms. In other words, high probability of accounting fraud will be present when (i) accounts receivables represent more than 29.2\% of total assets; (ii) the proportion of current assets in relation to total assets is lower than 0.347; and (iii) total liabilities are 13.2\% or higher than shareholders' equity.

\vskip 2mm
\textit{Transportation, Communication, Electric, Gas and Sanitary Service}
\vskip 0.5mm

Moderate accuracy performance is achieved by the proposed methods in this case, being boosted trees and random forests the ones showing the best results. On the one hand, random forests perform well when predicting fraud cases (75\%), as opposed to boosted trees that perform better when predicting non-fraud cases (71.9\%).

\begin{figure}[H]
\caption{Decision Tree Visualisation \\ Industry: Transp., Comm., Electric, Gas and Sanitary Serv.}
%\caption{Decision Tree Visualisation \\ Industry: Transportation, Communication, Electric, \\ Gas and Sanitary Service}
\centering
\label{DT_SIC4}
\includegraphics[scale=0.45]{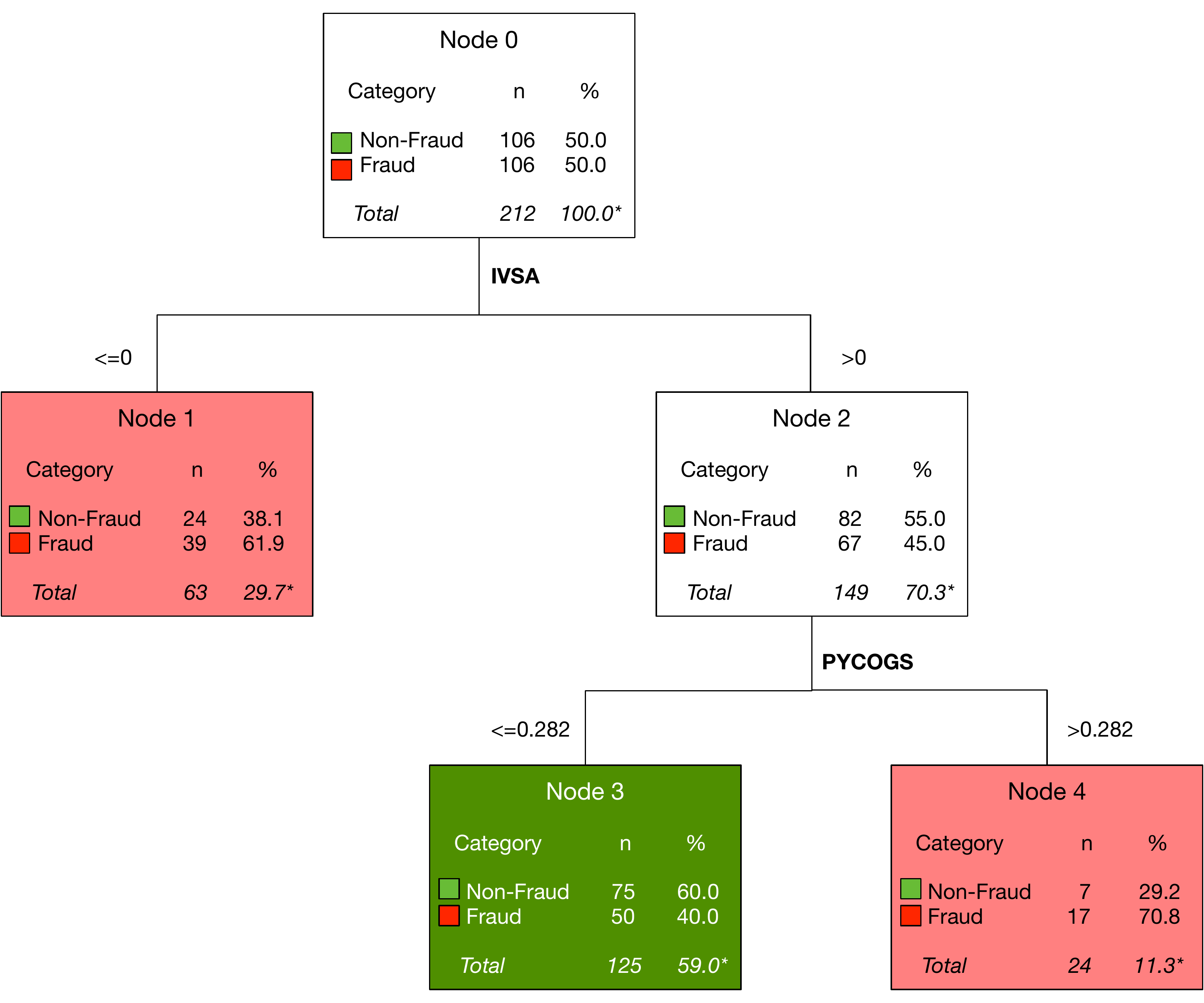}
\end{figure}

More relevant results can be observed from Figure \ref{DT_SIC4}. The most significant predictors of accounting fraud committed in this industry are IVSA and PYCOGS. As such, fraudulent reporting is more likely to be occurring as a result of misstatement of inventory levels and/or accounts payable figures.

As for the case of inventory manipulation, the warning sign is triggered when IVSA is lower or equal than zero. From basic accounting, it is known that figures of inventory and total sales cannot be negative due to the lack of economic meaning. Then, the only possibility in this case is that inventories are zero. Consequently, auditors should be cautious when null inventories are part of financial statements as it may be a sign of accounting fraud.

On the other hand, if inventory levels are not null, then fraud alarm should be activated when accounts payable represent 28.2\% of cost of good sold, as it may be indicating fraudulent activities of firms belonging to the industry at issue.

\newpage

\vskip 2mm
\textit{Wholesale Trade and Retail Trade}
\vskip 0.5mm

Moderate accuracy is achieved in the case of trading firms. It can be observed that boosted trees show superior performance when detecting both fraud and non-fraud cases. Decision trees, on the other hand, achieved exceptional results when predicting fraud instances, but poor performance when dealing with non-fraud cases.

\begin{landscape}
\begin{figure}
\caption{Decision Tree Visualisation \\ Industry: Wholesale Trade and Retail Trade}
\label{DT_SIC5}
\centering
\includegraphics[scale=0.45]{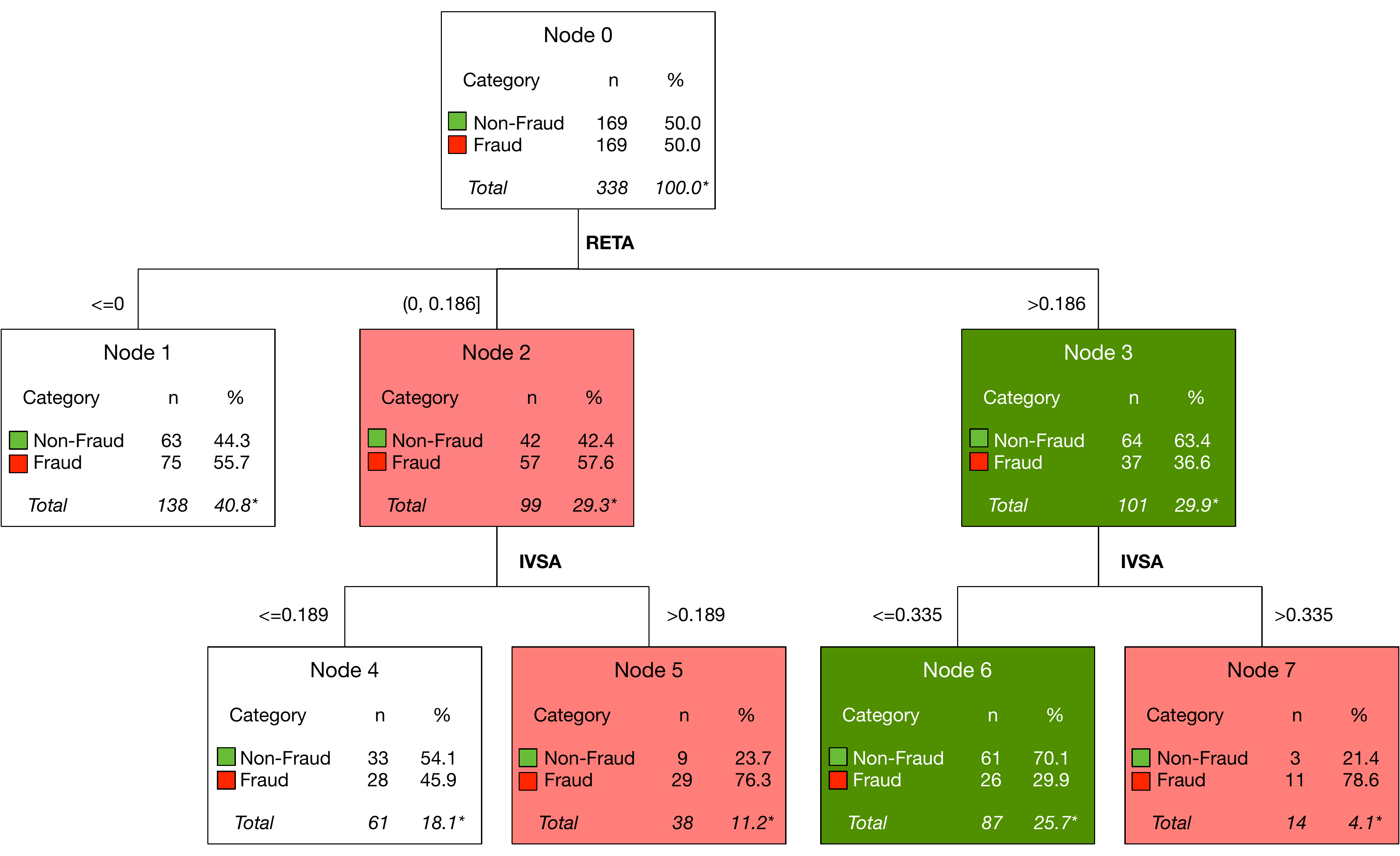}
\end{figure}
\end{landscape}

Furthermore, decision trees results suggest that fraudulent trading companies manipulate mainly two financial items simultaneously, that is, retained earnings and inventories. Two clear patterns can be identified when accounting fraud is being committed, as shown in Figure \ref{DT_SIC5}.

The first pattern has been found when the RETA ratio is between 0 and 0.186, and the IVSA ratio is higher than 0.189. That is, moderate positive values of retained earnings and large values of inventory happening together represents a clear sign of falsified reports.

The second pattern of fraudulent activity is identified when the RETA ratio is higher than 0.186 and, at the same time, the IVSA ratio is higher than 0.335 That is, exaggerated valuation of earnings compared to assets, and inventory compared to sales are considered in this industry as irregular, hence more attention should be paid when facing this situation.

\vskip 2mm
\textit{Finance, Insurance and Real Estate}
\vskip 0.5mm

Moderate prediction accuracy is obtained again, now in the industry of Finance, Insurance and Real Estate. In general, more advanced models achieved slightly better performance, out of which boosting techniques perform the best. In particular, it can be seen that boosted trees correctly classify 68.2\% of non-fraud cases and 63.2\% of fraud cases.

\begin{landscape}
\begin{figure}
\caption{Decision Tree Visualisation \\ Industry: Finance, Insurance and Real Estate}
\label{DT_SIC6}
\centering
\includegraphics[scale=0.45]{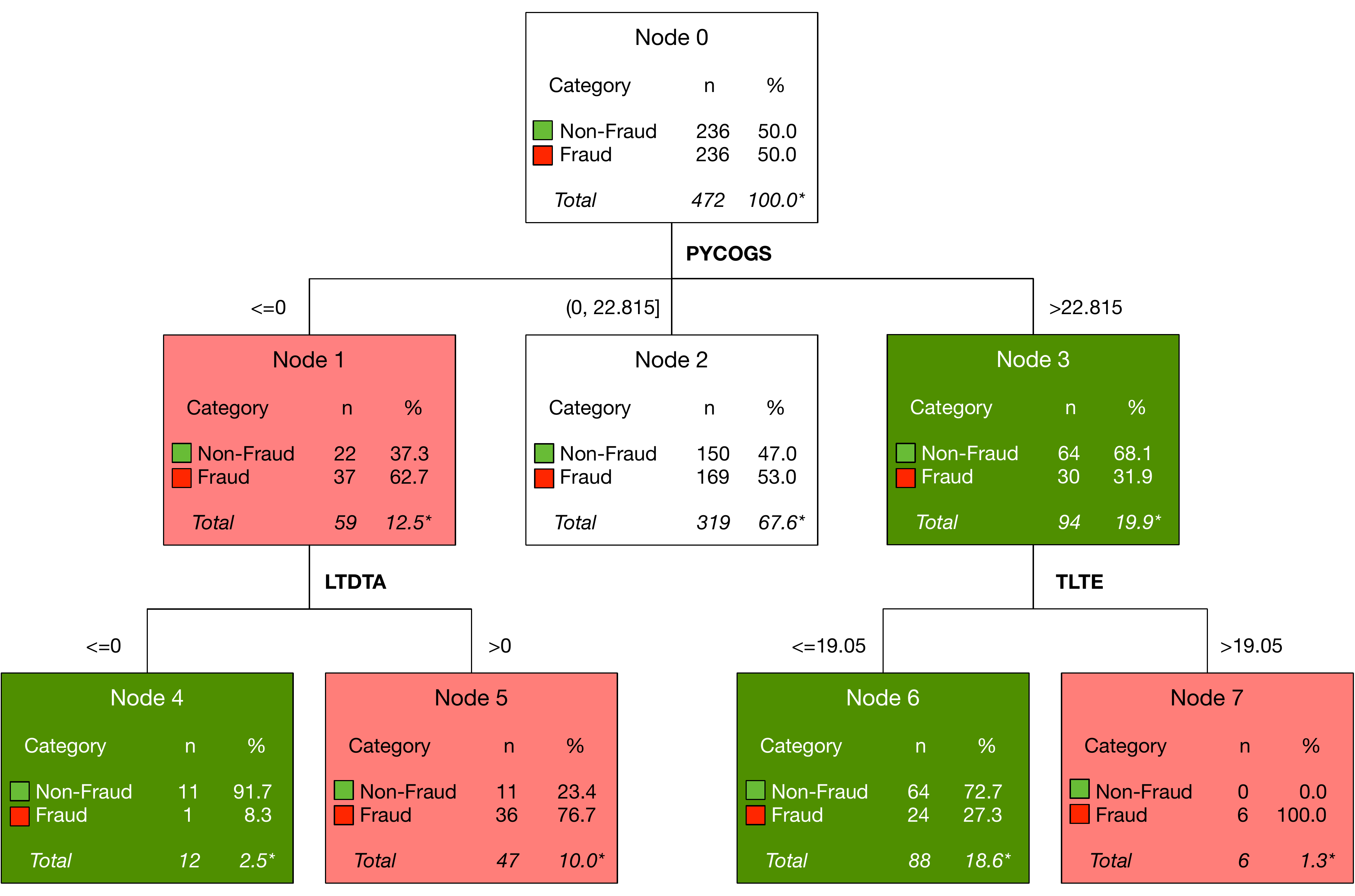}
\end{figure}
\end{landscape}

Moreover, and as it can be seen in Figure \ref{DT_SIC6}, fraudulent reporting within financial firms is more likely to be occurring as a result of manipulation of accounts payable and debt-specific figures. On the one hand, if accounts payable are lower or equal to zero together with long-term debt higher than zero, then more attention must be paid as it may be a sign of accounting fraud.

On the other hand, if accounts payable to cost of good sold are higher than 22.82 and, simultaneously, total liabilities are 19.05 times more than shareholders' equity, then warning alarm should be activated as irregular patterns are occurring that suggest fraudulent activities.

\vskip 2mm
\textit{Services}
\vskip 0.5mm

Poor performance achieved by machine learning methods when detecting accounting fraud within the service industry. Relatively better performance attained by tree-based methods, being decision trees the methodology that showed a more balanced performance regarding correct positive and negative classifications, that is, between sensitivity and specificity.

\begin{figure}[H]
\caption{Decision Tree Visualisation \\ Industry: Services}
\label{DT_SIC7}
\centering
\includegraphics[scale=0.45]{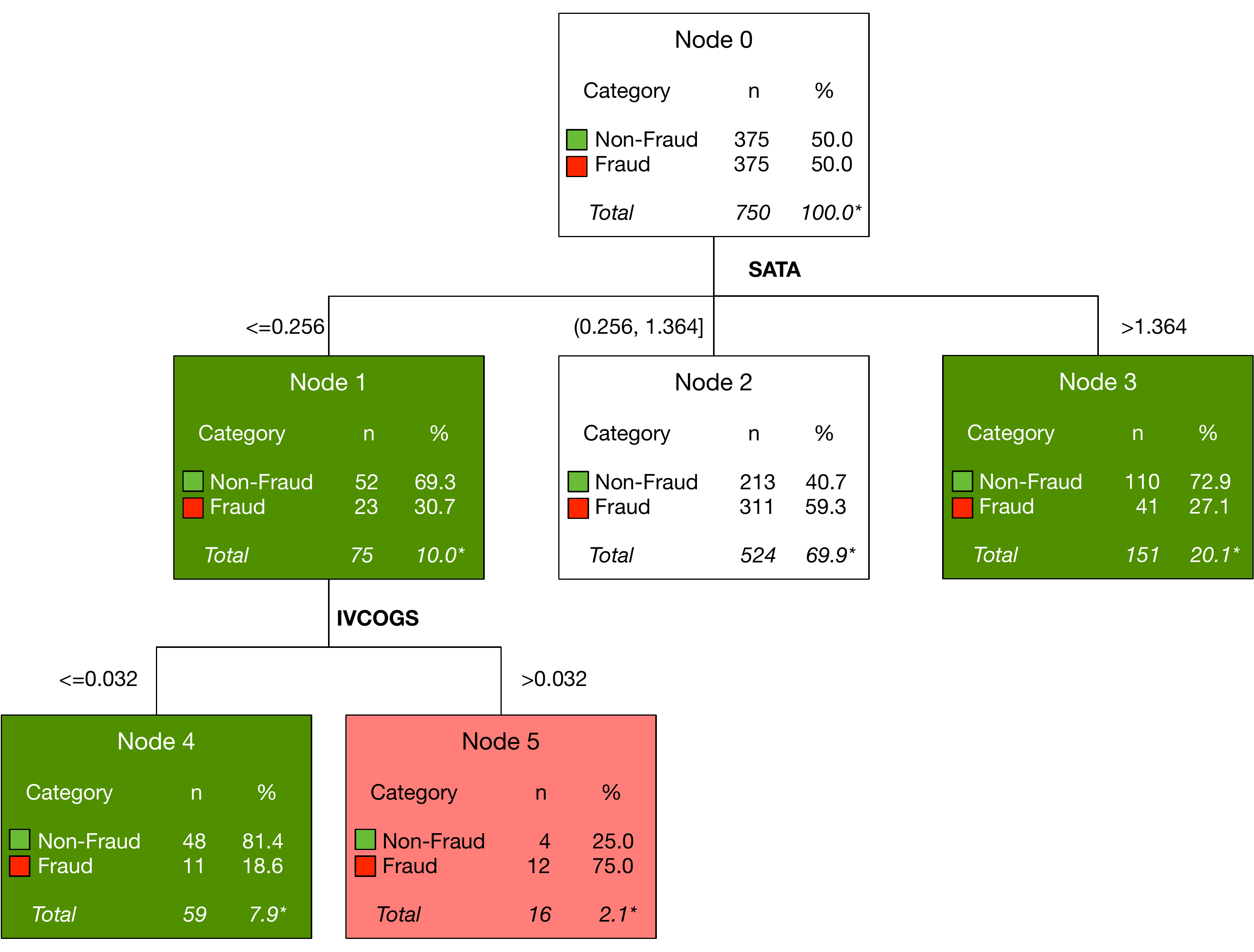}
\end{figure}

In addition, and as depicted in Figure \ref{DT_SIC7}, a fairly straightforward trick is usually performed by fraudulent companies in the industry of service, that is understating of sales figure together with the artificial exaggeration of inventory. More scrutiny should be made when total sales represent less than 25.6\% of total assets, as well as when the proportion of inventory in terms of cost of good sold is higher than 0.032, as they may be indicating that accounting fraud is being conducted.

\vskip 2mm
\textit{Public Administration}
\vskip 0.5mm

Exceptional results are obtained in the industry of public administration. Particularly superior performance was accomplished by random forests, as 90\% of non-fraudulent cases are correctly classified, as well as 83.8\% of fraudulent cases.

\begin{figure}[H]
\caption{Decision Tree Visualisation \\ Industry: Public Administration}
\label{DT_SIC8}
\centering
\includegraphics[scale=0.45]{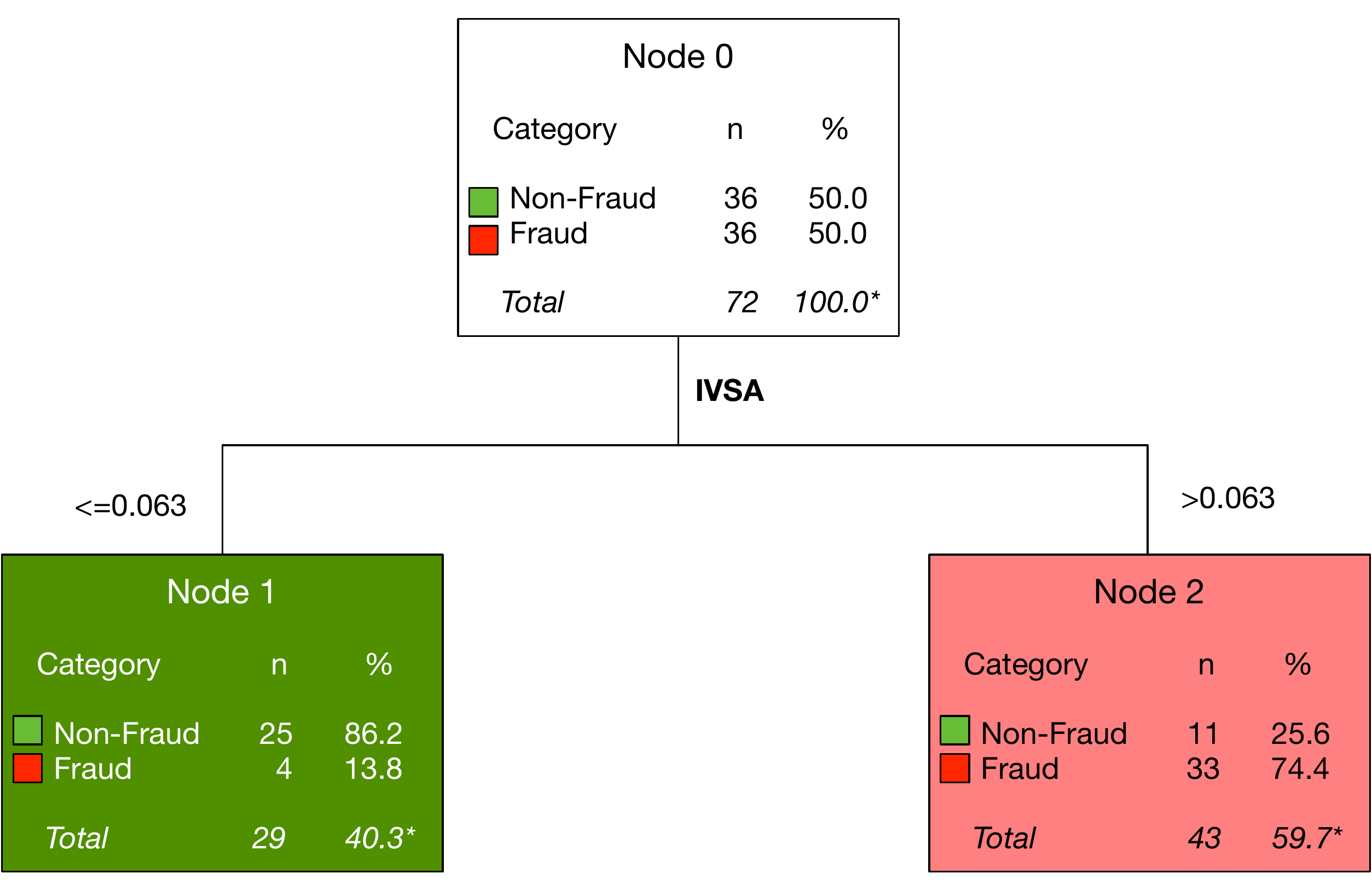}
\end{figure}

Accounting fraud in the industry of public administration is highly related to large values of inventory compared to sales, as it can be seen in Figure \ref{DT_SIC8}. Furthermore, special attention should be paid when evidencing inventories representing 6.3\% or more of total sales, as this is a clear sign of manipulated financial reports.

%
%More specifically, favourable average accuracy, specificity and sensitivity rates in the industry of Finance, Insurance and Real Estate are shown. These findings also suggest that it is easier to detect firms with a higher potential of committing fraud than not committing fraud.
%
%In brief, results suggest that all machine learning methods proposed in this paper provide superior predictive power compared to a naive strategy of classifying all firms as one class either fraud or non-fraud.

\newpage
\section{Conclusions}
\label{conclusions}

\subsection{Conclusions}

This study aims to identify signs of accounting fraud occurrence to be used to, first, identify companies that are more likely to be manipulating financial statement reports, and second, assist the task of examination within the riskier firms by evaluating relevant financial red-flags, as to efficiently recognise irregular accounting malpractices.

To achieve this, a thorough forensic data analytic approach is proposed that includes all pertinent steps of a data-driven methodology. First, data collection and preparation is required to present pertinent information related to fraud offences and financial statements. Then, an in-depth financial ratio analysis is performed in order to analyse the collected data and to preserve only meaningful variables. Finally, statistical modelling of fraudulent and non-fraudulent instances is performed by implementing several machine learning methods, followed by the extraction of distinctive fraud-risk indicators related to each economic sector.

This study contributes in the improvement of accounting fraud detection in several ways, including the collection of a comprehensive sample of fraud and non-fraud firms concerning all financial industries, an extensive analysis of financial information and significant differences between genuine and fraudulent reporting, selection of relevant predictors of accounting fraud, contingent analytical modelling for better differentiate between non-fraud and fraud cases, and identification of industry-specific indicators of falsified records.

The results of the current research suggest there is a great potential in detecting falsified accounting records through statistical modelling and analysis of publicly available accounting information. It has been shown good performance of basic models used as benchmark - discriminant analysis and logistic regression-, and better performance of more advanced methods, including AdaBoost, decision trees, boosted trees and random forests. Results support the usefulness of machine learning models as they appropriately meet the criteria of accuracy, interpretability and cost-efficiency required for a successful detection system.

The proposed methodology can be easily used by public auditors and regulatory agencies in order to assess the likelihood of accounting fraud, and also to be adopted in combination with the experience and knowledge of experts to lead to better examination of accounting reports. In addition, the proposed methodological framework could be of assistance to many other interested parties, including investors, creditors, financial and economic analysts, amongst others.

\subsection{Limitations and Future Work}
\label{limitationsandfuturework}

The collected sample of accounting fraud offences is considered to be only a fragment of the population of companies issuing fraudulent financial statement, as there is no guarantee that non-fraudulent firms are in fact legitimate observations until proven otherwise. Also, non-public companies are excluded from this study as the SEC only has jurisdiction over publicly traded companies. 

It is worth noting that accounting fraud is very versatile, and as such, will always evolve in terms of deceptive tricks. Managers will adapt their fraudulent schemes in order to successfully commit fraud, hence results obtained in this study are exclusively consequence of the investigation of the collected data and different conclusions may be reach when considering an alternative source of information.

Lastly, models performances are not ideal in some scenarios mainly due to sample size and omitted predictive variables. It is strongly suggested the inclusion of additional information to help better understand the accounting fraud phenomenon, which may consist of qualitative variables, including corporate governance information and inside trading data, as well as time-evolving features and industry-trending benchmarks. It would not be surprising to discover interesting temporal patterns of stock prices or asset returns when dealing with fraudulent corporations, or find an extraordinary economic performance of dishonest companies compared to the industry average.

Further work can be done for classification threshold selection. When modelling the accounting fraud phenomenon, it was mentioned that a specific classification threshold was considered to determine fraud and non-fraud categories in several machine learning techniques. Evaluation of different thresholds would be of much interest as it may improve classification accuracy in a cost-sensitive environment, such as the one at issue.

In addition, different methodologies are suggested to tackle the imbalance class challenge. The method adopted in the present study was based on random under-sampling, but other techniques may improve this part of the process, such as random over-sampling, bootstrap models, cost modifying methods and algorithm-level approaches, to name a few.

More advanced machine learning techniques are also recommended. It would be very interesting to implement alternative and more advanced methods, such as support vector machines, neural networks and Bayesian models, as they may be helpful to correctly identify fraudulent firms.

Finally, it is suggested to replicate the proposed methodology in specific economic domains, such as the pharmaceutical industry, health care industry and financial industry, amongst others. The more specialised the industry, the more interesting patterns are likely to be found and, therefore, to be explored and analysed.

\section*{Acknowledgements}
\label{acknowledgements}

The authors would like to thank the Securities Class Action Clearinghouse, Stanford Law School, for providing access to the collection of fraud cases considered in this study.

\newpage
\section*{References}
\label{references}

\bibliographystyle{abbrvnat}
\bibliography{phd_literature}

\footnotesize
\begin{itemize}
\item[\authorimg{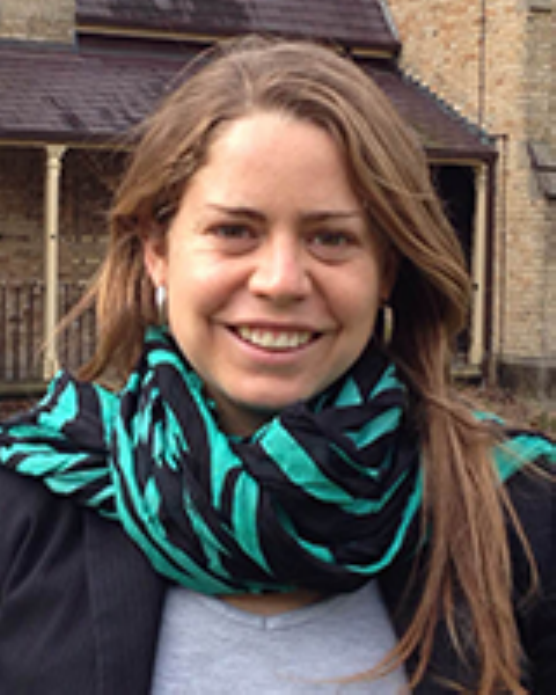}] \begin{minipage}{\dimexpr\linewidth-2\baselineskip\relax}
{\bf Maria Jofre} is a Casual Lecturer at the Discipline of Business Analytics, The University of Sydney Business School. She holds a B. Eng and a MS from the University of Chile, and a Ph.D. from the University of Sydney. Her areas of expertise and interest are statistical machine learning and corporate criminology.
                                    \end{minipage}
\item[\authorimg{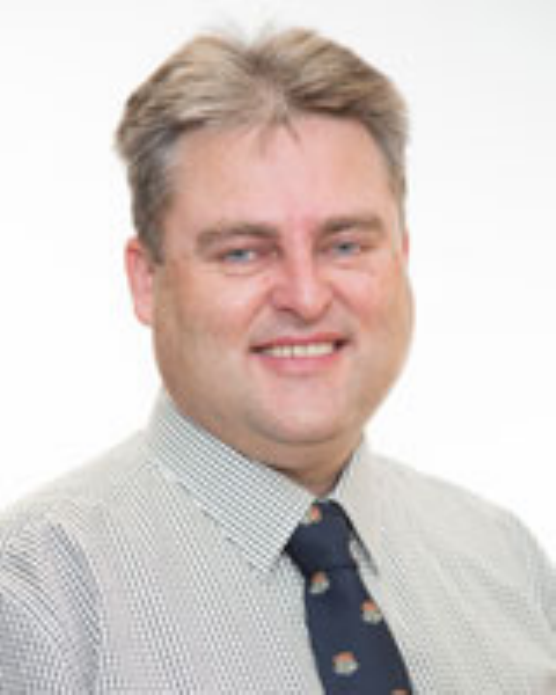}] {\bf Richard Gerlach}'s research interests lie mainly in financial econometrics and time series. His work has concerned developing time series models for measuring, forecasting and managing risk in financial markets as well as computationally intensive Bayesian methods for inference, diagnosis, forecasting and model comparison. Recent focus has been Value-at-Risk and Expected Shortfall forecasting. He has developed structural break and intervention detection tools for use in state space models; also has an interest in estimating logit models incorporating misclassification and variable selection. His applied work has involved forecasting risk levels during and after the Global Financial Crisis; assessing asymmetry in major international stock markets, in response to local and exogenous factors; stock selection for financial investment using logit models; and option pricing and hedging involving barriers.

His research papers have been published in Journal of the American Statistical Association, Journal of Business and Economic Statistics, Journal of Applied Econometrics, Journal of Time Series Analysis and the International Journal of Forecasting. He has been an invited speaker and regular presenter at international conferences such as the International conference for Computational and Financial Econometrics, the International Symposium on Forecasting and the International
\end{itemize}

\end{document}